\pdfoutput=1
\documentclass{article}

\usepackage[preprint]{main}

\usepackage[utf8]{inputenc} % allow utf-8 input
\usepackage[T1]{fontenc}    % use 8-bit T1 fonts      % hyperlinks
\usepackage{url}            % simple URL typesetting
\usepackage{booktabs}       % professional-quality tables
\usepackage{amsfonts}       % blackboard math symbols
\usepackage{nicefrac}       % compact symbols for 1/2, etc.
\usepackage{microtype}      % microtypography
\usepackage{xcolor}         % colors
\usepackage{amsmath}
\usepackage{amssymb}
\usepackage{nomencl}
\usepackage{glossaries}
\usepackage{xspace}
\usepackage{multirow}
\usepackage{graphicx}
\usepackage{bbding}
\usepackage{pifont}
\usepackage{algorithm}
\usepackage{graphicx}
\usepackage{subfigure}
\usepackage{enumitem}
\usepackage{multicol}
\usepackage{caption}
\usepackage{array}
\usepackage{fp}
\usepackage{makecell}
\usepackage{flushend}
\usepackage{cases}
\usepackage{color}
\usepackage{algpseudocode}
\usepackage{listings}
\usepackage{mathrsfs}
\usepackage{longtable}
\usepackage{colortbl}
\usepackage{bm}
\usepackage{tabularx}
\usepackage{natbib}
\usepackage{caption}
\usepackage{subcaption}
\usepackage{soul}
\usepackage{graphicx} % 插入图像
\usepackage{subcaption} % 子图支持
\usepackage{hyperref}
\setlist[itemize]{noitemsep, topsep=0pt}
\definecolor{codegreen}{rgb}{0,0.3,0.6}
\definecolor{codegray}{rgb}{0.5,0.5,0.5}

\newcommand{\ie}{\emph{i.e.,}\xspace}
\newcommand{\eg}{\emph{e.g.,}\xspace}

\newcommand{\ignore}[1]{}

\usepackage{tcolorbox}
\definecolor{darkorange}{RGB}{255, 140, 0}
\definecolor{lightgreen}{RGB}{145, 204, 117}
\definecolor{lightyellow}{RGB}{250, 200, 88}
\definecolor{lightred}{RGB}{238, 102, 102}
\definecolor{lightblue}{RGB}{115, 192, 222}
\newtcolorbox{promptbox}[2][Prompt]{
colback=black!5!white,
arc=5pt, 
boxrule=0.5pt,
fonttitle=\bfseries,
title=#1, 
before upper={\scriptsize}, fontupper=\fontfamily{ptm}\selectfont,
colframe=#2, % 使用传递的参数来设定 colframe
}

\title{From Trial-and-Error to Improvement: A Systematic Analysis of LLM Exploration Mechanisms in RLVR}

\author{
  Jia Deng\thanks{Equal contribution.} ,~Jie Chen\textsuperscript{*},
  \textbf{Zhipeng Chen,~Daixuan Cheng,}\\
  \textbf{Fei Bai,~Beichen Zhang,~Yinqian Min,~Yanzipeng Gao,~Wayne Xin Zhao\thanks{Correspondence to Wayne Xin Zhao.}~,~Ji-Rong Wen} \\
  \vspace{1pt} \\
  Gaoling School of Artificial Intelligence, Renmin University of China\\
  \small{\texttt{dengjia0510@outlook.com, ptyzchenjie@ruc.edu.cn, batmanfly@gmail.com}}
}

\begin{document}
\textit{Technical Report on Slow Thinking with LLMs: Exploration Mechanism}
\maketitle
\begin{abstract}
Reinforcement learning with verifiable rewards (RLVR) has emerged as a powerful paradigm for enhancing the reasoning capabilities of large language models (LLMs). Unlike traditional RL approaches, RLVR leverages rule-based feedback to guide LLMs in generating and refining complex reasoning chains—a process critically dependent on effective exploration strategies. While prior work has demonstrated RLVR’s empirical success, the fundamental mechanisms governing LLMs’ exploration behaviors remain underexplored.

This technical report presents a systematic investigation of exploration capacities in RLVR, covering four main aspects: (1) exploration space shaping, where we develop quantitative metrics to characterize LLMs’ capability boundaries;
(2) entropy-performance exchange, analyzed across training stages, individual instances, and token-level patterns; and (3) RL performance optimization, examining methods to effectively translate exploration gains into measurable improvements. 
By unifying previously identified insights with new empirical evidence, this work aims to provide a foundational framework for advancing RLVR systems. We release our resources at the STILL project website:  \url{https://github.com/RUCAIBox/Slow_Thinking_with_LLMs}.\footnote{This work is an extension of our previous work, ``\emph{Decomposing the Entropy-Performance Exchange: The Missing Keys to Unlocking Effective Reinforcement Learning~\cite{deng2025decomposing}}''.}.
\end{abstract}

\section{Introduction}
\label{sec-intro}
Recently, reinforcement learning (RL) has significantly enhanced the complex reasoning abilities of large language models (LLMs)~\cite{zhao2023survey}. A key distinction from early RL approaches (\eg RLHF~\cite{rlhf} with trained reward models) for training LLMs is the incorporation of \emph{verifiable rewards}. 
The underlying idea is simple, yet the effect can be surprisingly powerful: when provided with verifiable signals, LLMs learn to produce lengthy reasoning chains that ultimately yield correct answers~\cite{deepseekr1}. A well-known precedent for this approach is AlphaGo~\cite{AlphaGo}, which used rule-based rewards to train Go models. This methodology is now widely recognized as \emph{reinforcement learning with verifiable rewards (RLVR)}~\cite{rlvr}.

In RLVR, LLMs first generate rollout responses to training problem prompts and then leverage these self-generated responses to improve model performance. This learning process iterates until performance gains become negligible. 
A crucial aspect of RLVR is enabling effective exploration within the vast state space of natural language. Research~\cite{reason_explore} has shown that the exploration capabilities of LLMs not only influence immediate learning progress but also determine the ultimate performance of the models. Thus, developing a systematic understanding of LLMs' exploration abilities—and how they drive performance improvements—is essential for RLVR.

To investigate the exploration mechanism in RLVR, we first revisit the fundamental exploration-exploitation trade-off in the classic RL literature~\cite{explorationsurvey}. An RL agent must strategically balance \emph{exploration} (testing novel actions to discover improved strategies) and \emph{exploitation} (leveraging known optimal actions to earn immediate rewards). This balance is crucial: excessive exploration delays convergence, while insufficient exploration may lead to locally optimal but globally subpar policies. 
In RLVR, verifiable rewards enable LLMs to guide their exploration in a task-aligned manner. The framework uses exploratory actions to identify potentially correct solutions to reasoning tasks, then reinforces successful solutions while pruning unsuccessful attempts—creating a self-improving cycle of reasoning refinement.

Given the pivotal role of exploration mechanisms in RLVR, this domain has drawn considerable research interest, spanning investigations of entropy mechanisms~\cite{entropymechanism,reason_explore} (where entropy reduction enhances performance) to various enhancement techniques~\cite{dapo,nsr} (\eg Clip-higher). 
However, despite these advances, current studies have predominantly examined either isolated or coarse-grained aspects of exploration mechanisms. A comprehensive understanding of several fundamental issues remains lacking, particularly regarding how to properly structure the exploration space, how exploration precisely translates to performance gains, and how to effectively augment exploration capabilities.

In this technical report, we conduct a systematic investigation of the fundamental exploration mechanisms employed by LLMs in RLVR. Our methodology integrates a comprehensive literature review with rigorous empirical analysis. The discussion is organized around three key dimensions:

$\bullet$ \emph{Exploration space structure} (Section~\ref{sec:explo}): We investigate methods to structure the exploration space for LLMs, with particular focus on developing quantitative metrics to characterize their capability boundaries. This involves determining both the solvable and unsolvable problems within practical LLM rollout constraints. Furthermore, we also compare how two primary post-training approaches—SFT and RL—influence LLM exploration capabilities and overall performance. 

%$\bullet$ \emph{Training method effects} (Section~\ref{}): We examine how two primary post-training approaches—supervised fine-tuning (SFT) and reinforcement learning (RL)—influence LLM exploration capabilities and overall performance. Our analysis includes a systematic comparison of how these methods structure the exploration space.

$\bullet$ \emph{Entropy-performance interplay} (Section~\ref{sec:inter}): 
We investigate the relationship between \emph{entropy} (a key indicator of exploration capability) and model performance. Our analysis extends beyond reviewing recent advances in this area to include a multi-granularity empirical examination across three levels: stage-level
dynamics, instance-level efficiency, and token-level significance. 

$\bullet$ \emph{Performance improvement} (Section~\ref{sec:performance improvement}): 
We discuss approaches to enhancing reasoning abilities, with a particular focus on two main aspects: (1) expanding exploration capacities and (2) enhancing the performance conversion efficacy. Concretely, we first review recent advancements in strengthening the exploration abilities of LLMs. Moreover, we conduct experiments to investigate how to preserve Pass@k performance during training and propose two simple methods to boost the RL efficiency.

Overall, this report establishes a foundational framework for understanding LLMs' exploration mechanisms in RLVR and their role in enhancing reinforcement learning performance. Through an integrated approach combining literature synthesis with novel empirical analysis, we offer a comprehensive investigation of these mechanisms and their practical implications. 

\section{Quantifying Exploration Abilities in RLVR} %(czp)
\label{sec:explo}

% how ability boundary relates to exploration 

% A simple paragraph for Definition of LLM Ability Boundary. 

%The exploration ability of an LLM refers to its capacity to solve problems through trial-and-error attempts~\cite{}.
In the context of RL, the exploration abilities of LLMs refer to their capacity to discover effective solutions through iterative trial-and-error and environmental interaction~\cite{math_shepherd}, particularly in complex reasoning or planning tasks. 
A model with stronger exploration ability can solve problems more efficiently, requiring fewer attempts. These successful attempts then serve as training data for improving the model's capabilities through RL algorithms like GRPO~\cite{deepseekmath}. 
The exploration process is fundamentally constrained by the model's \emph{ability boundary}, \ie the upper limit of its problem-solving capacity~\cite{ability_boundary}. For RLVR training to be effective, the LLM must generate successful attempts within this explorable space; problems beyond this boundary cannot be adequately solved through RL optimization.
Therefore, we propose to quantify LLMs' exploration capabilities by measuring their ability boundaries. Specifically, this section introduces two key metrics for this assessment. 
%Prior work~\cite{} evaluates exploration ability using the Pass@k metric and the entropy of the policy distribution.
%Given the conceptual link between exploration ability and ability boundary, these metrics can be adapted to estimate a model's ability boundary as well.

%If, after sufficient exploration, the model still fails to solve a problem, we conclude that the problem lies beyond its ability boundary—the upper limit of its problem-solving potential~\cite{}.

%Both exploration ability and ability boundary depend on whether the model can eventually solve a problem given enough attempts.
%If the model succeeds through repeated trials, the problem falls within its ability boundary; otherwise, it exceeds the model's current capacity.

%2.1和2.2放到一起，主要两大类：passk和entropy

\subsection{Pass@k Metric}
%k取到一个比较合理的数值，比如k=100

The Pass@k metric, widely adopted in prior studies~\cite{prorl,reason_explore,nsr}, evaluates whether LLMs can solve problems within $k$ attempts. It has been adopted to estimate models' ability boundaries~\cite{passk,ability_boundary,passk_training}.

In this part, we first formalize the Pass@k metric and its unbiased estimator. We then present the extensions for adapting Pass@k to assess LLMs' exploration abilities.

%Pass@k is a commonly used metric to measure whether LLMs can solve the problems in $k$ attempts~\cite{}, which is used to measure the ability boundary in prior studies~\cite{}. 
%In this part, we first introduce the formulation of the Pass@k metric and its unbiased estimation, and then provide the approach to utilize the Pass@k metric to assess the ability boundary of LLMs.

\paragraph{Formulation for Pass@k Metric.} 
Given a question $q$, the LLM generates $k$ responses denoted by $\{o_i\}_{i=1}^{k}$, each evaluated by a verifier that assigns a binary reward: $1$ for correct responses and $0$ for incorrect ones. The Pass@k score for the problem equals $1$ if at least one response is correct, and $0$ otherwise. Formally, the Pass@k metric is defined as: 
\begin{equation}
\small
    \text{Pass@k}=\mathbb{E}_{q\sim\mathcal{D},\{o_i\}_{i=1}^{k}\sim\pi_{\theta}(\cdot|q)}\left[\max\left(R(o_1),\dots,R(o_k)\right) \right],
\end{equation}
where $\pi_\theta$ refers to the policy model with parameters $\theta$, and $R(o_i)$ denotes the reward of $i$-th response generated by policy model.
However, the choice of $k$ significantly impacts the stability of Pass@k calculation, with small $k$ values leading to high variance. To mitigate this, prior work~\cite{passk} introduced an unbiased estimator for Pass@k. 
%To address the issue, previous work~\cite{} proposes the unbiased estimation of the Pass@k metric.
Concretely, the model first generates the $n$ responses $\{o_i\}_{i=1}^{n}$ based on the given question $q$ ($n\geq k$), and then Pass@k metric is computed by the expection of whether $k$ responses contain the positive one among $\{o_i\}_{i=1}^{n}$. 
This leads to the following unbiased estimation: 
\begin{equation}
    \small
    \text{Pass@k}=\mathbb{E}_{q\sim\mathcal{D},\{o_i\}_{i=1}^{n}\sim\pi_{\theta}(\cdot|q)}\left[1-\frac{\binom{n-c}{k}}{\binom{n}{k}}\right],
\end{equation}
where $c$ is the number of the positive responses among the sampled responses $\{o_i\}_{i=1}^{n}$.
Since the unbiased estimation of Pass@k provides stable evaluation of model capabilities, we adopt this method for computing Pass@k in our subsequent analysis.

%As the unbiased estimation of Pass@k can bring a stable evaluation of model ability, we adapt it for computing the Pass@k metric in the following discussion.

% \paragraph{$k$-rollout Solvable Problems.} By using the Pass@k metric, we can estimate a model's exploration abilities by sampling a number of responses and checking for correct solutions. Our empirical observations reveal that when the sample size reaches a large value, additional sampling rarely yields new correct solutions. 
% %This motivates us to define \emph{$k$-rollout Solvable Problems}, which counts the number of problems solvable within $k$ attempts ($k=xxx$ in our work) as a pratical indicator of the model's ability boundary.
% This motivates our \emph{$k$-rollout Solvable Problems} metric, which counts problems solvable within $k$ attempts (\textcolor{blue}{$k=640$ in our implementation}) as a practical measure of the model's ability boundary. 
% With this definition, when comparing models, we can use Venn diagrams to visualize the overlaps among 
% $k$-rollout solvable problems.
% Models with superior ability boundaries would demonstrate both larger solution sets and greater coverage of other models' solvable problems. 
\paragraph{$k$-rollout Unsolvable Problems.}
The Pass@k metric allows us to identify the limits of a model's exploration abilities by sampling numerous responses and determining which problems consistently lack a correct solution. Our empirical observations reveal that once the sample size becomes sufficiently large, additional sampling rarely leads to new correct solutions, effectively stabilizing the set of problems the model cannot solve. This motivates our \emph{$k$-rollout Unsolvable Problems} metric, which identifies the set of problems that remain unsolvable after $k$ attempts as a practical measure of a model's capability limitations. With this definition, when comparing models, we can use Venn diagrams to visualize the overlaps among their sets of $k$-rollout unsolvable problems. By analyzing the overlap and relative sizes of the sets of unsolvable problems, we can identify which models possess a higher capability boundary (\ie a smaller set of unsolvable problems).

%To visualize comparisons between models, we employ Venn diagrams showing the overlap in their $k$-rollout solvable problems. Models with broader ability boundaries demonstrate: (1) larger total solvable problem sets, and (2) greater coverage of other models' solvable problems.

% Individual measurement: Number of Solvable Problems 
% Pairwise comparison: Venn diagram
%Inspired by the Pass@k metric, through sampling a large number of responses generated by the model, judging whether the sampled responses contain correct solutions can be used to indicate the capability boundary of the model.
%In practice, we observe that when the sampling number $k$ reaches 1024, it becomes difficult for the model to solve new problems even with more sampling.
%Thus, the number of solvable problems within $k=1024$ attempts (named as \emph{$k$-rollout Solvable Problems}) can be regarded as an individual measurement to assess the LLM ability boundary.
%Based on this idea, the Venn diagram can be used to compare the ability boundary of different LLMs.
%Concretely, we compute which problem can be solved in $k$ rollouts for each model.
%Through analyzing the overlap of $k$-rollout solvable problems, models with higher capability boundaries can be identified.

\subsection{Entropy of Policy Distribution}
\label{sec:ent}
In addition to the Pass@k metric, the entropy of policy distribution serves as another important measure for assessing LLM exploration abilities~\cite{entropymechanism}. This metric captures the diversity and uncertainty in LLM behaviors during problem-solving.
In this section, we first formalize the token-level entropy computation, then introduce an extended metric—the rollout branching factor—to quantify LLM exploration capabilities.

\paragraph{Formulation of Token-level Entropy.}
In the context of LLMs, token-level entropy quantifies the uncertainty in the model's token generation process. Following standard information theory definitions~\cite{entropy_information_theory}, the entropy $H_i$ of the $i$-th token is computed as:
\begin{equation}
    H_i=-\sum_{t_i\in \mathcal{V}} \pi_\theta(t_i|t_{<i})\log{\pi_\theta(t_i|t_{<i})},
    \label{eq:token_entropy}
\end{equation}

where $\mathcal{V}$ and $t_{<j}$ denote the model vocabulary and the previously generated tokens, respectively.
This formulation implies that for a given problem, higher values of $H_i$ indicate greater uncertainty in the LLM's generation of the $i$-th token, reflecting stronger exploratory behavior at that token position.
%According to this formulation, given a problem, the higher value of $H_i$ refers to higher uncertainty of LLMs in generating the $i$-th token, suggesting that LLMs can perform better exploration during generating the $i$-th token.

\paragraph{Rollout Branching Factor.}
The uncertainty of a model's token generation can be measured by the number of candidate tokens with relatively high probabilities, which we term the \emph{Rollout Branching Factor}.  Following common decoding hyperparameters ($\text{top-p}=0.95$), we consider all tokens within the top 95\% probability mass as potential candidates for generation. 
We define these as the candidate tokens. Crucially, a larger number of candidate tokens indicates greater generation diversity for the LLM, corresponding to stronger exploration ability and a higher ability boundary.

%The uncertainty of a model in generating tokens can be reflected by the number of candidate tokens with relatively high probabilities, named as \emph{Rollout Branching Factor}.
%As the commonly used hyperparameter settings in the decoding process, \ie $\text{top\_p}=0.95$, tokens whose sum of probabilities is less than or equal to $0.95$ may be sampled and used as the final generated tokens.
%Thus, we regard these tokens as the candidate tokens.Building on this, a higher number of candidate tokens shows that LLMs have more options to perform generation, relating to higher exploration ability and ability boundary.

\subsection{Extended Discussion}

The Pass@k metric and policy distribution entropy represent two widely-used methods for evaluating LLM ability boundaries. Previous work~\cite{nsr,entropymechanism} has additionally demonstrated that the entropy of LLM-generated responses reflects the model's problem-solving uncertainty, where greater uncertainty may indicate higher potential solution diversity and thus greater problem-solving capacity~\cite{reason_explore}. 
Furthermore, generalization performance on unseen tasks serves as another effective indicator - models with higher ability boundaries typically exhibit stronger zero-shot task transfer capabilities. 

However, a major issues with Pass@k is that it may yield correct final answers in RLVR without producing valid reasoning chains. This phenomenon has been empirically verified through manual inspection in previous work~\cite{verify_step_by_step}. Several studies have attempted to address this limitation by developing modified methods that verify the correctness of reasoning chains in addition to final answers. 
% \textcolor{blue}{add some discussions how do we manage that}
A potential approach is to use a language model to evaluate the reasoning chains and assess whether any errors are contained~\cite{genrm}.
Specifically, two levels of evaluation granularity can be conducted for LLM assesssment, \ie  process-level evaluation~\cite{qwen_prm} and instance-level evaluation~\cite{still_1}.
To better evaluate the reasoning chains, existing studies~\cite{genrm_deepseek,genrm_seed} employ reasoning models as evaluators, making the evaluation results more precise and reliable.

Currently, we do not employ any specific methods to mitigate this type of error. We conducted manual inspections of a random sample of outputs to verify the accuracy of the reasoning chains, finding that most correct answers are generally supported by valid reasoning rather than chance—this pattern is particularly evident for larger values of $k$. This observation aligns with findings reported in previous research~\cite{ability_boundary}. That said, we recognize that investigating the integrity and faithfulness of reasoning chains generated by language models constitutes an important research direction, which we reserve for future work.

%Currently, we don't adopt any specific methods to avoid this type of error. We manually inspected a random sample of outputs to confirm the correctness of the reasoning chains, and find that most of the correct answer are are generally supported by valid reasoning rather than chance, especially at larger values of $k$. This finding is consistent with that reported in the prior study~\cite{ability_boundary}. However, we believe that it is an important research topic to study the integrity and faithfulness of reasoning chains generatated by LLMs. We leave this topic as future work. 
 
%However, studies~\cite{ability_boundary} show that accurate final answers are generally supported by valid reasoning rather than chance, especially at larger values of $k$, preserving the integrity of Pass@k-based conclusions. Moreover, to further ensure the validity of our findings, we manually inspected a random sample of outputs to confirm the correctness of the reasoning chains, ensuring a faithful reflection of the models’ reasoning capabilities.

%  \textcolor{blue}{how do we deal with that}
% \textcolor{red}{we do not solve this issue in our previous experiments.}

\subsection{Comparison of SFT and RL Effects} 
\begin{figure}[htbp]
    \centering % 整个 figure 居中
    
    % 2. 使用 \subfigure 命令，注意语法变化
    \subfigure[Unbiased estimation of pass@k.]{ % 标题在方括号里
        \includegraphics[width=0.43\textwidth]{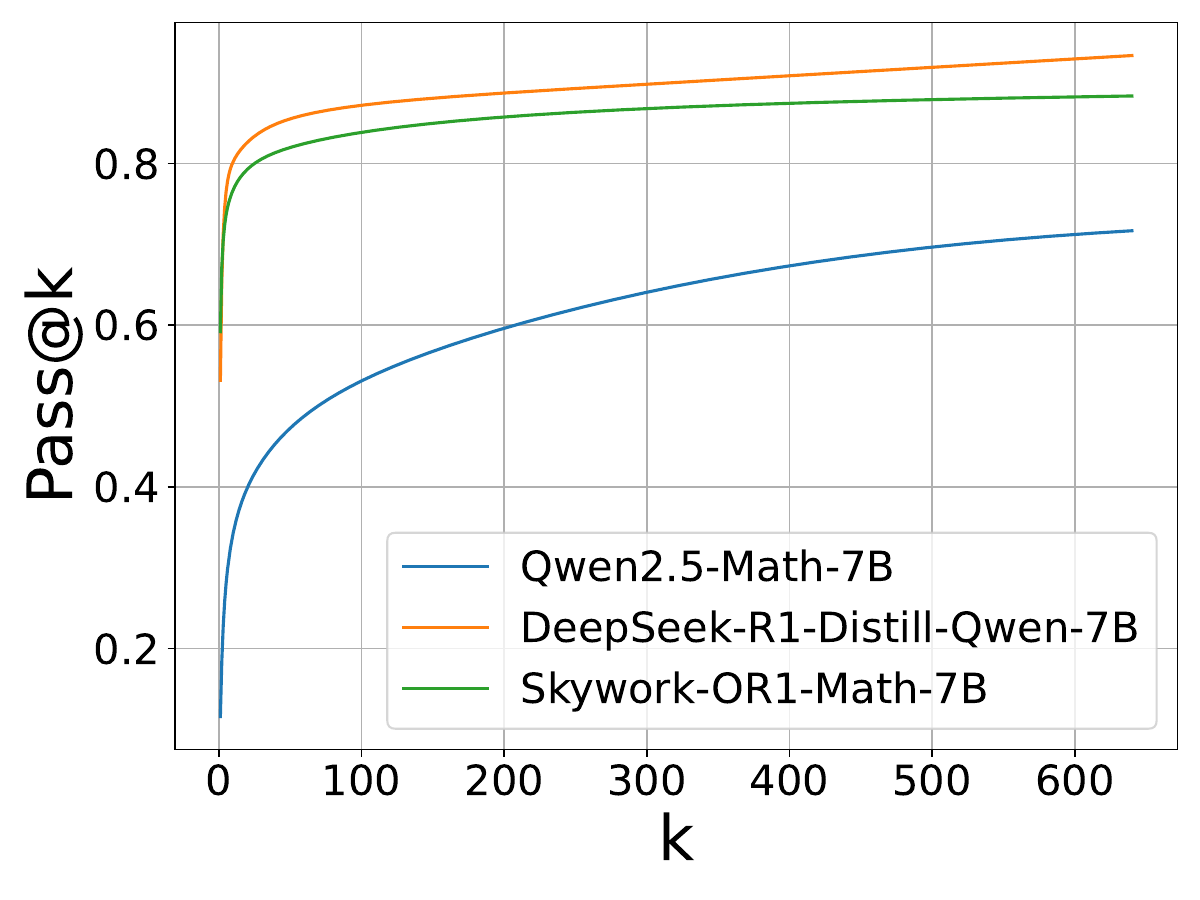}
        \label{fig:passk2.4} % 标签在花括号里
    }
    \hfill % 间距控制方式不变
    \subfigure[Comparison of answer diversity.]{
        \includegraphics[width=0.43\textwidth]{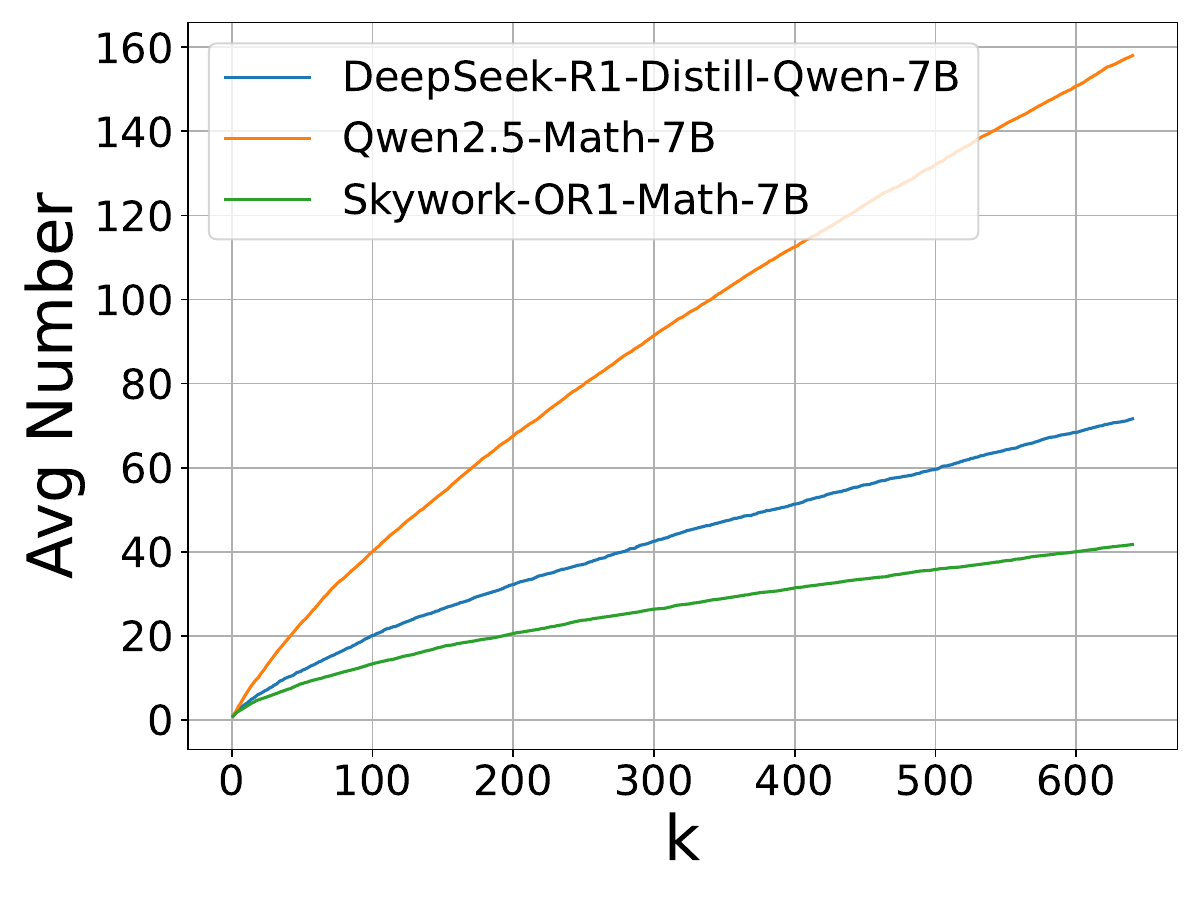}
        \label{fig:unique_answers_plot}
    }
    \hfill
    \subfigure[Venn diagram of unsolvable problems.]{
        \includegraphics[width=0.43\textwidth]{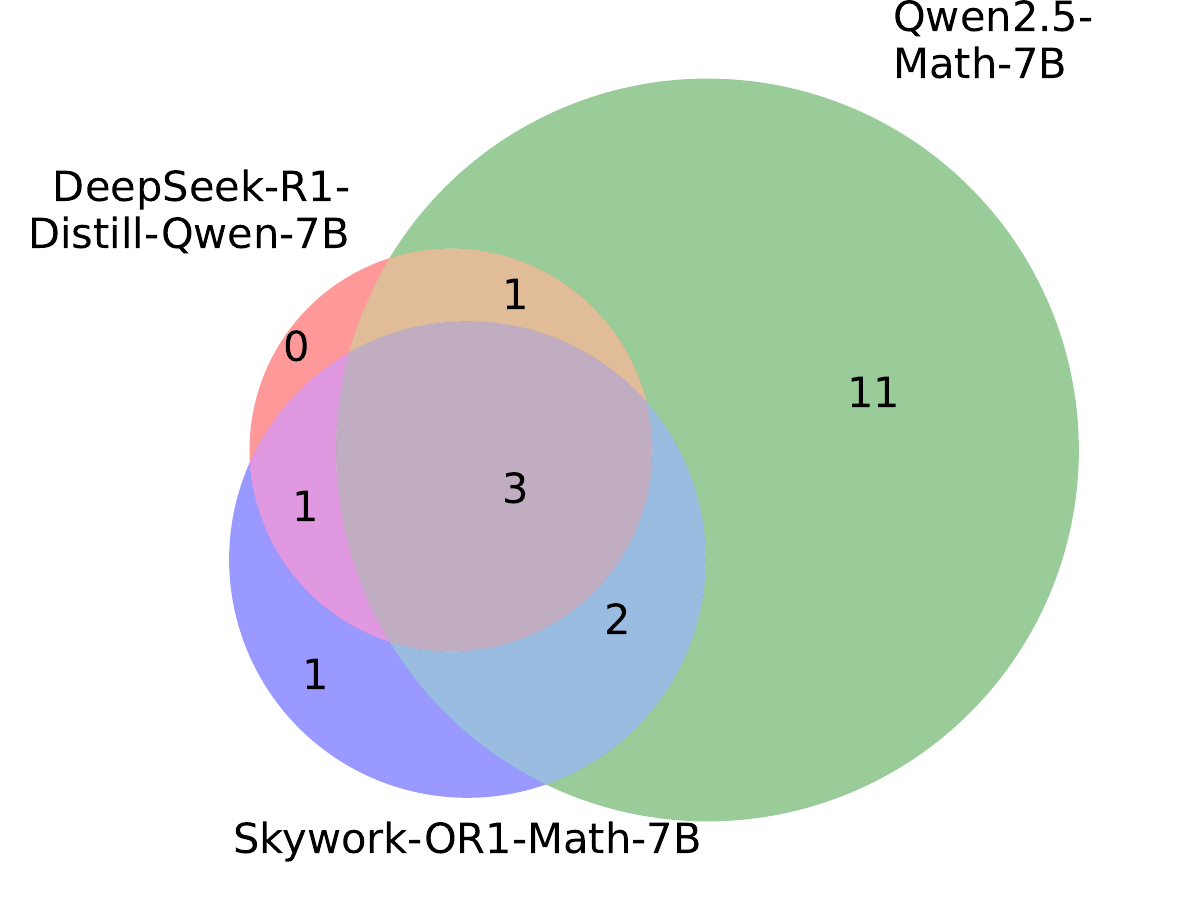}
        \label{fig:venn_ids}
    }
    \hfill
    \subfigure[Comparison of rollout branching factor.]{
        \includegraphics[width=0.43\textwidth]{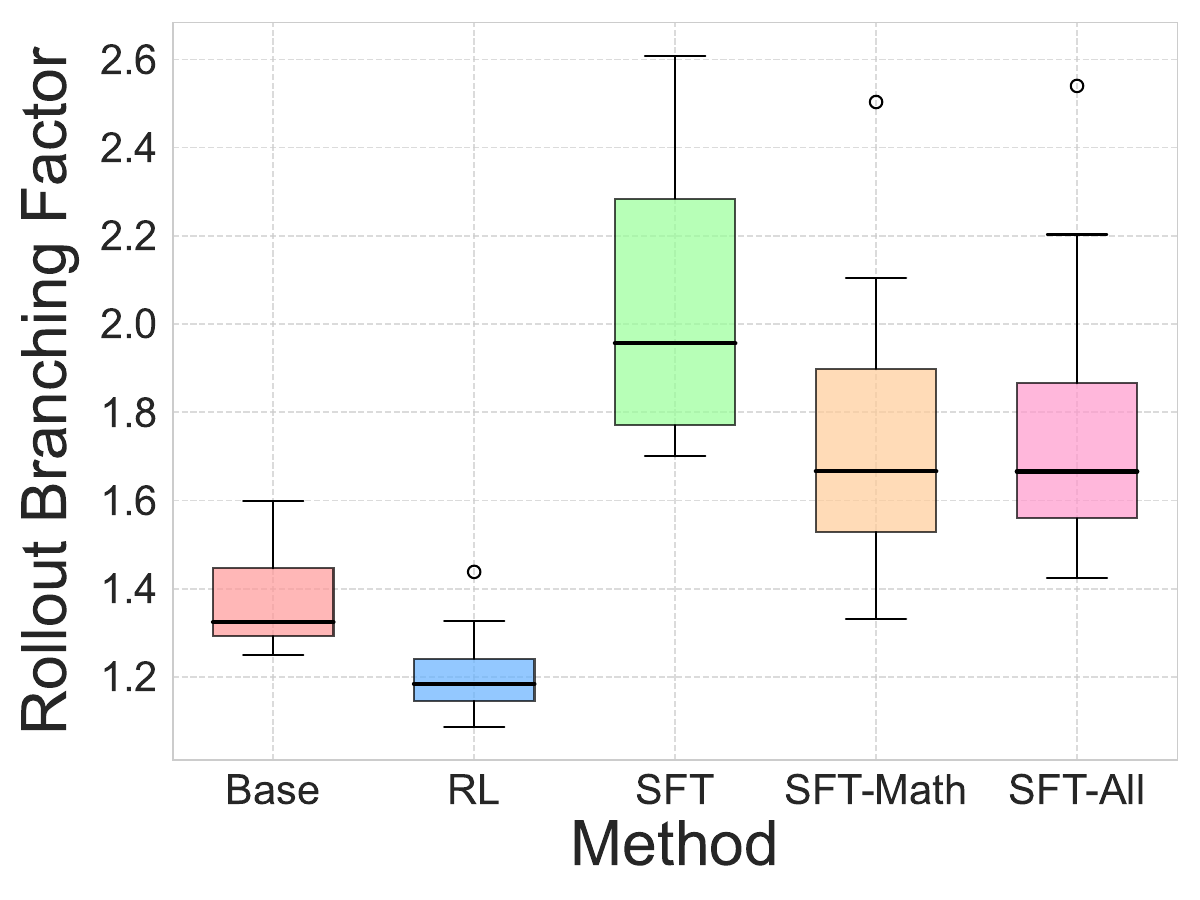}
        \label{fig:nucleus_size}
    }

    \caption{Comparison of different models' exploration capabilities across various metrics.} % 总标题命令不变
    \label{fig:2.1}
\end{figure}

% Our analysis investigates the effects of SFT and RL by comparing three models. We use Qwen2.5-Math-7B as our base model. This model is first enhanced using SFT to yield DeepSeek-R1-Distill-Qwen-7B. Subsequently, this SFT model is further trained with RL to produce Skywork-OR1-Math-7B.
\paragraph{Experimental Setup.}
Our analysis investigates the effects of SFT and RL by comparing three models selected to represent a sequential training pipeline: Qwen2.5-Math-7B serves as the base model; DeepSeek-R1-Distill-Qwen-7B is an SFT-enhanced version of the base; and Skywork-OR1-Math-7B is a version of the SFT model that has been further trained with RL. All evaluations are performed on $60$ questions from AIME24 and AIME25. For the analysis of the rollout branching factor, we also introduce two additional SFT models derived from the base model using the OpenThoughts dataset: SFT-Math, trained only on the math domain, and SFT-All, trained on all domains (\ie code, math, science, and puzzle). The SFT process for these models used a maximum context length of $20,000$ tokens, a batch size of $96$, and a learning rate of $1 \times 10^{-5}$. All experiments are conducted on a challenging set of 60 problems from AIME24 and AIME25, using $k=640$ attempts. We evaluate the models using Pass@k, answer diversity, k-rollout unsolvable problems, and the rollout branching factor. We set the temperature to $0.6$, the top-p to $0.95$, and allow a maximum generation of $8192$ tokens.

\paragraph{Results and Analysis.}
Our investigation compares SFT and RL across three key dimensions: their impact on overall problem-solving success, their effect on the capability boundary, and their influence on token-level generation diversity.

$\bullet$ \textbf{SFT expands the Pass@k boundary while RL sharpens Pass@1 performance.} We first compare how SFT and RL affect problem-solving ability and answer diversity. The experiment contrasts the base, SFT, and RL models on the AIME dataset. As shown in Figure~\ref{fig:passk2.4}, SFT achieves a significant improvement in model's Pass@k score compared to the base model (Qwen2.5-Math-7B \emph{v.s.} DeepSeek-R1-Distill-Qwen-7B). In contrast, RL shows no improvement in Pass@k, indicating a decline in the model's exploration capability (DeepSeek-R1-Distill-Qwen-7B \emph{v.s.} Skywork-OR1-Math-7B). However, RL is known to improve Pass@1, sharpening the model's ability to solve problems on the first attempt. This focus on exploitation comes at a cost: Figure~\ref{fig:unique_answers_plot} shows that both training methods reduce answer diversity, but the RL-trained model exhibits a much more pronounced decrease. This suggests that SFT effectively expands the model's capability boundary by learning from diverse external solutions, whereas RL optimizes and reinforces existing solution paths, leading to a more deterministic policy that improves exploitation (Pass@1) but suppresses exploration (Pass@k and answer diversity).

$\bullet$ \textbf{Both SFT and RL shift the capability boundary.} To gain a more granular understanding beyond aggregate metrics, we analyze how the set of solvable problems changes after training. We compare the sets of $k$-rollout unsolvable problems for the three 7B models using the same AIME24 and AIME25 test set. The Venn diagram in Figure~\ref{fig:venn_ids} reveals no complete set inclusion. This demonstrates that both SFT and RL induce bidirectional shifts in capability: problems solvable by the base model can become unsolvable after training, and vice-versa. This confirms that training does not simply expand the ability boundary but induces a shift in it. Notably, the RL model's set of unsolvable problems is larger than that of the SFT model. This suggests that RL is more prone to narrowing the model's exploration space.

$\bullet$ \textbf{SFT fosters token-level diversity, whereas RL leads to a more constrained policy.} Finally, we investigate how training impacts exploration at the token level by measuring the rollout branching factor. We compare the base model against the RL model (Skywork-OR1-Math-7B) and three SFT variants (DeepSeek-R1-Distill-Qwen-7B, SFT-Math, and SFT-All). The results in Figure~\ref{fig:nucleus_size} show that all SFT models, whether trained on a single domain (SFT-Math) or multiple domains (SFT-All, which is trained on code, math, science, and puzzle), significantly increase the rollout branching factor over the base model. This shows they generate a more diverse set of candidate tokens at each step. In contrast, the RL model fails to increase the branching factor. This indicates that access to high-quality external data via SFT is a critical mechanism for enhancing a model's intrinsic exploration capacity. RL, constrained by its self-generated data, reinforces existing high-probability pathways and does not foster the token-level diversity needed for broader exploration.

\subsection{Effects of External Tools}
% 1. 给标题加一个脚注标记
% \subsection[Effects of External Tools]{Effects of External Tools\footnotemark}
% % 2. 脚注内容放在标题后面
% \footnotetext{\ExternalToolsFootnote}

RL directs model improvement through reward signals but relies primarily on self-generated data, restricting capacity expansion. In contrast, tool-integrated reasoning typically involves invoking external tools—such as code interpreters—to execute model-generated content, a process fundamentally distinct from the reasoning patterns of standard text-based language models. We therefore hypothesize that integrating external tools may significantly extend a model’s reasoning capabilities. To test this hypothesis, we conducted an empirical study\footnote{This part is adapted from our earlier technical report, ``\emph{Towards Effective Code-Integrated Reasoning\cite{bai2025towards}}''. It is included here to ensure the completeness of the discussion regarding ability boundaries.}.

\paragraph{Environmental Setup.}
To investigate how tool-integrated reasoning(TIR) influences a model’s capability boundaries, we evaluated four baseline models—all derived from Qwen2.5-Math-7B—on the AIME2024 and AIME2025 benchmarks using the Pass@k metric. For all experiments, we set the temperature to 0.6, the top-p value to 0.95, and allowed a maximum generation of 16,384 tokens. The four baselines are as follows:

$\bullet$ \emph{Base$_{text}$}: The base model, utilizing standard text-based reasoning.

$\bullet$ \emph{Base$_{code}$}: The base model, utilizing code-integrated reasoning via the direct prompt.

$\bullet$ \emph{RL$_{text}$}: The model trained with vanila RL, utilizing text-based reasoning. It is trained in a standard RLVR setting with GRPO. 

$\bullet$ \emph{RL$_{code}$}: The model trained with tool-augmented RL, utilizing code-integrated reasoning.
It is trained using an approach that integrates a code interpreter; refer to the paper~\cite{bai2025towards} for details. 

\paragraph{Results and Analysis.}

\begin{figure}[htbp]
\centering
\resizebox{0.9\linewidth}{!}{ % 缩小到 90% 宽度
\begin{tabular}{cc}
\subfigure[AIME2024]{\label{fig:passk_24}
\includegraphics[width=0.43\linewidth]{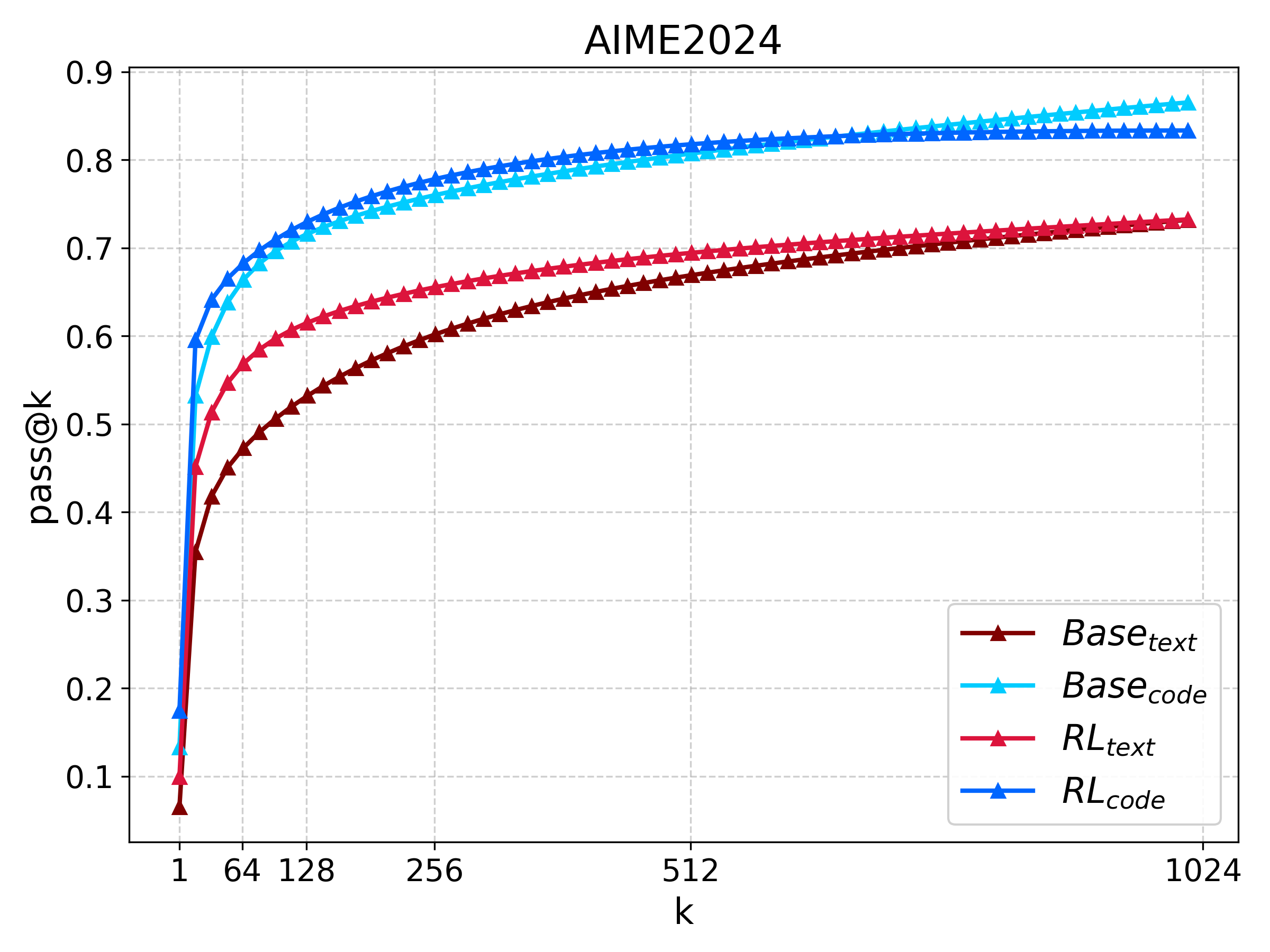}} &
\subfigure[AIME2025]{\label{fig:passk_25}
\includegraphics[width=0.43\linewidth]{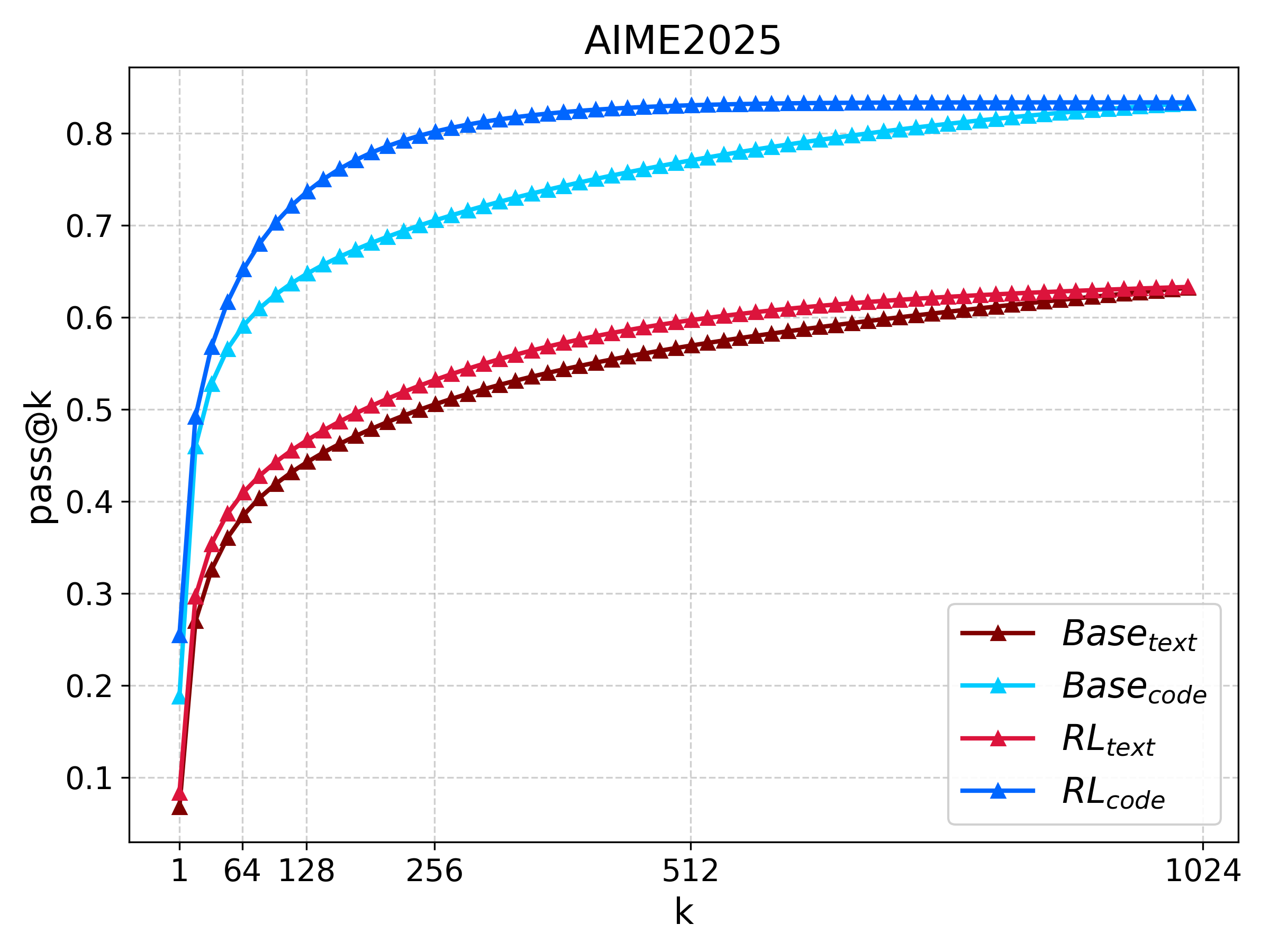}}
\end{tabular}
}
\caption{Pass@k Accuracy on AIME2024 and AIME2025.}
\label{fig:passk}
\end{figure}

% $\bullet$ \textbf{TIR expands the exploration capability boundary.}
As shown in Figure \ref{fig:passk}, TIR enhances Pass@k performance relative to standard text-based reasoning. This demonstrates that TIR can expand a model’s capability boundary by facilitating external computation and structured tool utilization. However, consistent with prior findings \cite{ability_boundary}, we observe that as $k$ increases, models trained with RL are eventually matched or even outperformed by their non-RL counterparts—a pattern that underscores the inherent limitations of RL in extending a model’s capability boundary.

\begin{tcolorbox}
[colback=black!5!white,colframe=black!55!white,title=\textbf{\footnotesize \textsc{\ul{Takeaway for Shaping LLM Ability Boundaries (Section 2)}}:}]
\begin{itemize}
[leftmargin=1mm]
\setlength\itemsep{0em}
    \item[\ding{224}] {\footnotesize 
    {\fontfamily{phv}\fontsize{8}{9}\selectfont
    SFT expands the model's exploration boundary by leveraging external data to increase token-level diversity and Pass@k performance, whereas RL sharpens exploitation for better Pass@1 at the cost of narrowing the exploration space and reducing answer diversity.}} 
    \item[\ding{224}] {\footnotesize 
    {\fontfamily{phv}\fontsize{8}{9}\selectfont
    Integrating external tools provides a powerful pathway to expand the ability boundary by facilitating external computation and structured tool utilization to significantly boost Pass@k, although this approach still suffers from the inherent limitations of RL in extending a model's capability boundary.}}
\end{itemize}
\end{tcolorbox}

%\input{sections/3.training_influence}
%\section{Fine-grained Analysis of LLM Abilities Improving}

\section{Understanding the Entropy-Performance Interplay}
\label{sec:inter}
From a broader perspective, exploration at the core of RLVR embodies a shift from \emph{uncertainty} to \emph{performance}. Initially, this uncertainty manifests as an expansive exploration space, enabling a diverse range of behavioral attempts. As exploration progresses, the model reinforces successful behavioral trajectories, gradually building confidence in these solutions—thereby driving performance improvements—while mitigating behavioral uncertainty. Ultimately, the model settles into a subtle trade-off between uncertainty and performance.
 
Specifically, building on the discussion in Section~\ref{sec:explo}, we adopt the entropy of the policy distribution as a metric for uncertainty and investigate the entropy-performance trade-off. We begin by reviewing existing literature on entropy-performance exchange mechanism, then present a fine-grained analysis supported by empirical experiments.

\subsection{Reviewing Existing Studies}
Recent advancements in RLVR for LLMs have underscored the critical role of entropy in facilitating effective exploratory reasoning. Overall, these studies yield two key insights into the entropy-performance relationship.

\paragraph{Dynamics of Entropy-Reward Exchange.}
Despite the complexity of the learning process, the underlying mechanism governing entropy-reward dynamics exhibits remarkable simplicity. Empirical results~\cite{entropymechanism} reveal an exponential relationship between performance and entropy, where the reward $R$ follows a predictable exponential decay with respect to policy entropy $H$. Formally, this relationship can be expressed as: 
\(
R = -a \cdot \exp(H) + b,
\)
where \(a, b > 0\) are constants that can be fit from empirical data. are constants that can be fit to empirical data. Using the fitted values of $a$
 and $b$, we can establish a macroscopic performance trend as a function of the varying entropy values. 
Specifically, $a$ governs the efficiency of coverting enttropy to performance, while $b-a$ defines the theoritical maximum reward achievable through RL training. 

\paragraph{Exploration Signals from High-Entropy Tokens.} 
As observed in \cite{reason_explore}, entropy dynamics are strongly linked to \textit{exploratory reasoning behaviors} in LLMs, with high-entropy tokens/sentences frequently serving as pivotal elements that guide or connect reasoning steps (\eg \texttt{first}, \texttt{because}, \texttt{however}), enable reflective actions like self-verification and correction, and facilitate rare reasoning strategies and under-explored behaviors by the base model. These findings suggest that entropy acts not only as a regularization signal of the RL algorithm but also as an intrinsic indicator of exploration capabilities. Besides, analysis in~\cite{beyond82rule} has revealed that only a minority of tokens exhibit high entropy, referred to as \textit{forking tokens}. These tokens serve as critical decision points where the model’s reasoning trajectory can diverge into multiple plausible paths. 

\subsection{Fine-grained Analysis} 

Despite these progresess, current investigations of the entropy-performance trade-off operate at a coarse granularity, treating RLVR training as a monolithic process or categorizing tokens into groups. These studies primarily examine aggregate performance changes before and after training states, failing to provide a fine-grained analysis of how entropy dynamics interact with model performance throughout the training trajectory. 
In essence, RLVR training constitutes a complex learning process shaped by multiple involving elements~\cite{entropymechanism}. These factors dynamically influence model behavior~\cite{wang2025stabilizing}, with entropy effects varying across training stages, token positions, and sampled instances—each contributing distinctively to overall performance.
Building on the above discussion, in this section, we conduct a systematic study of the entropy-performance interplay in RLVR, focusing on three key aspects: stage-level dynamics, instance-level efficiency, and token-level significance. 

\begin{figure}[t]
  \centering

  % 图1：Entropy and accuracy 曲线图
  \subfigure[Entropy and accuracy trends on AIME24, AIME25 and MATH500 during GRPO training. The red line marks the transition from the rising stage to the plateau stage.\label{fig:ent_and_acc}]{
    \includegraphics[width=0.43\textwidth]{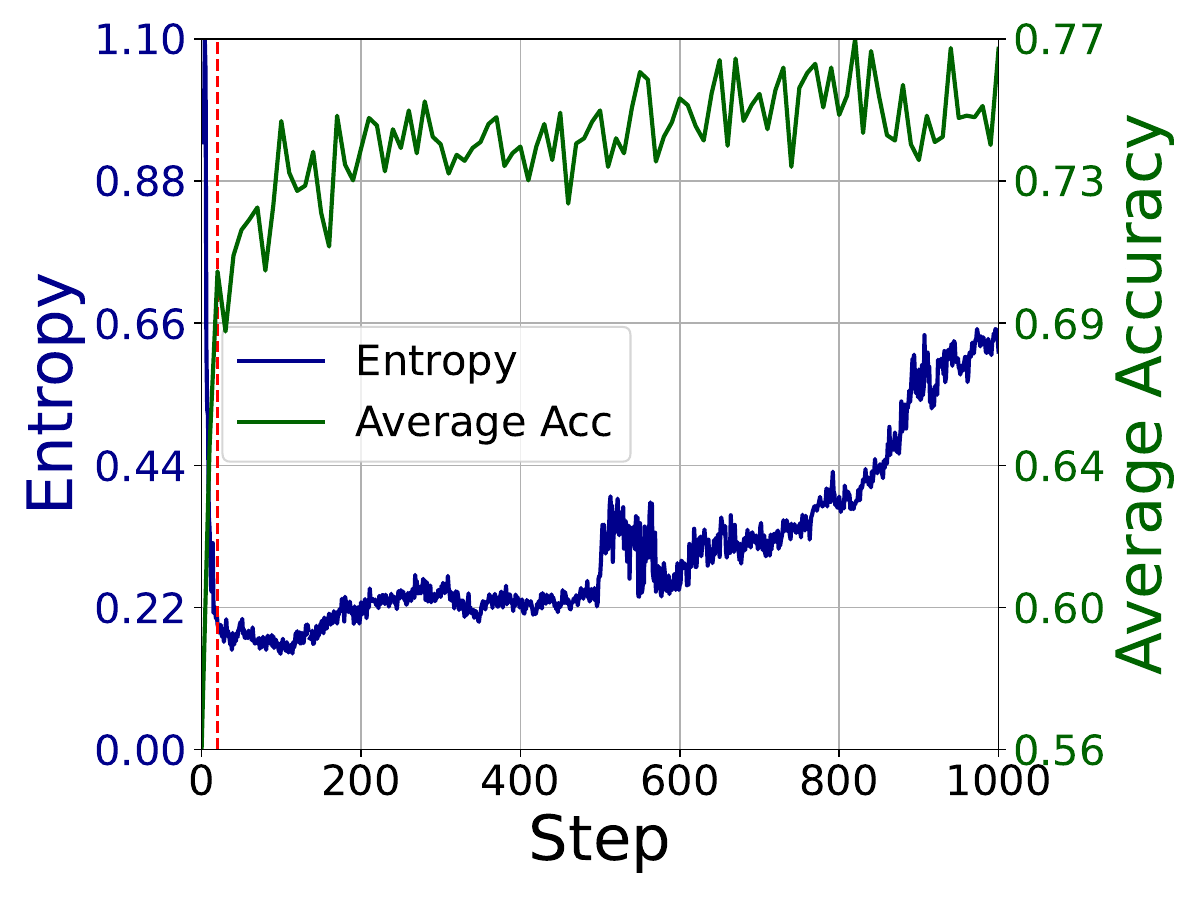}
  }
  \hspace{0.001\textwidth}
  % 图2：Token 表格
  \subfigure[Example of generated tokens with significant entropy decrease or frequency increase during the rising stage.\label{tab:token-table}]{
  \raisebox{24mm}{ % 抬高整个盒子,值可以微调
    \small
    \begin{minipage}[b]
    {0.43\textwidth}
    \vspace{-40mm}
      \centering
      \begin{tabular}{p{\dimexpr 0.5\linewidth-2\tabcolsep\relax}|p{\dimexpr 0.5\linewidth-2\tabcolsep\relax}}
        \hline
        \textbf{Entropy $\downarrow$} & \textbf{Frequency $\uparrow$} \\
        \hline
        Û, erot, whim, Dead,ther, flirt,ĉquery,erot, whim, Dead, ther, flirt, ĉquery, pornography, @SuppressWarnings, Ethics, Aad, UDA, è©, åĩĢåĢ¼,  ...&
        \textbackslash, \textbackslash\textbackslash, (, 2, 1, =, \{, -, +, frac, \}, \}\{, 3,5, ),\_, \textbackslash\}\textbackslash, x, 7, \textbackslash[, )., sqrt, times,  cdot,**, 9, (\textbackslash, 8, :, 5, ), 6,[, 0, 4, \textbackslash), ], left, right, k, y, n, pi, theta, )\textbackslash\textbackslash, \}\textbackslash\textbackslash, \{x, c, k, \_n, mod, cos, ):, )\^, )), f, sin, number, is, ...\\
        \hline
      \end{tabular}
    \end{minipage}
  }
  }
  \hspace{0.001\textwidth}
  % 图3：Entropy ratio 条形图
  \subfigure[Entropy dynamics. The bar chart shows the sample distribution of the top 20\% tokens exhibiting the fastest entropy drop.\label{fig:pos_isu}]{
    \includegraphics[width=0.43\textwidth]{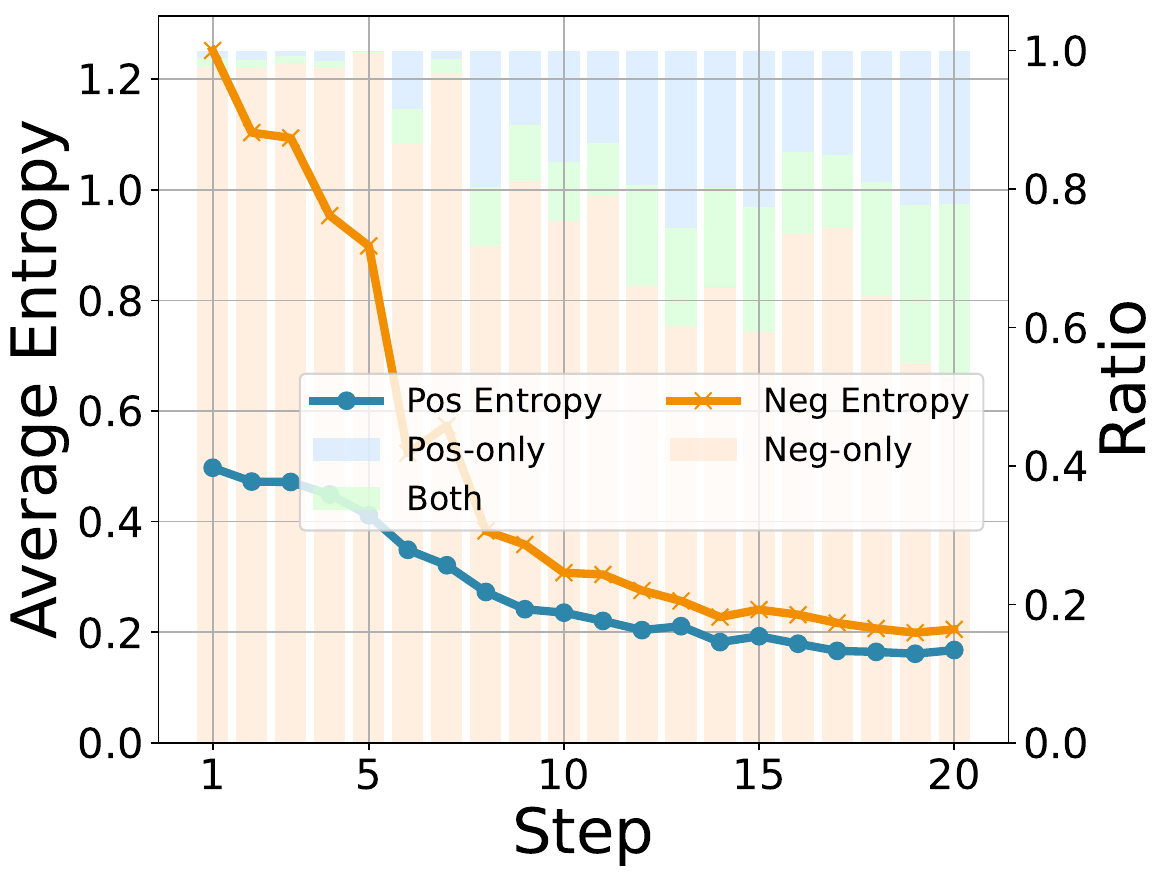}
  }
  \hspace{0.001\textwidth}
  % 图4：负面问题分布图
  \subfigure[Proportion of model responses containing quality issues across different training steps.\label{fig:isu}]{
    \includegraphics[width=0.43\textwidth]{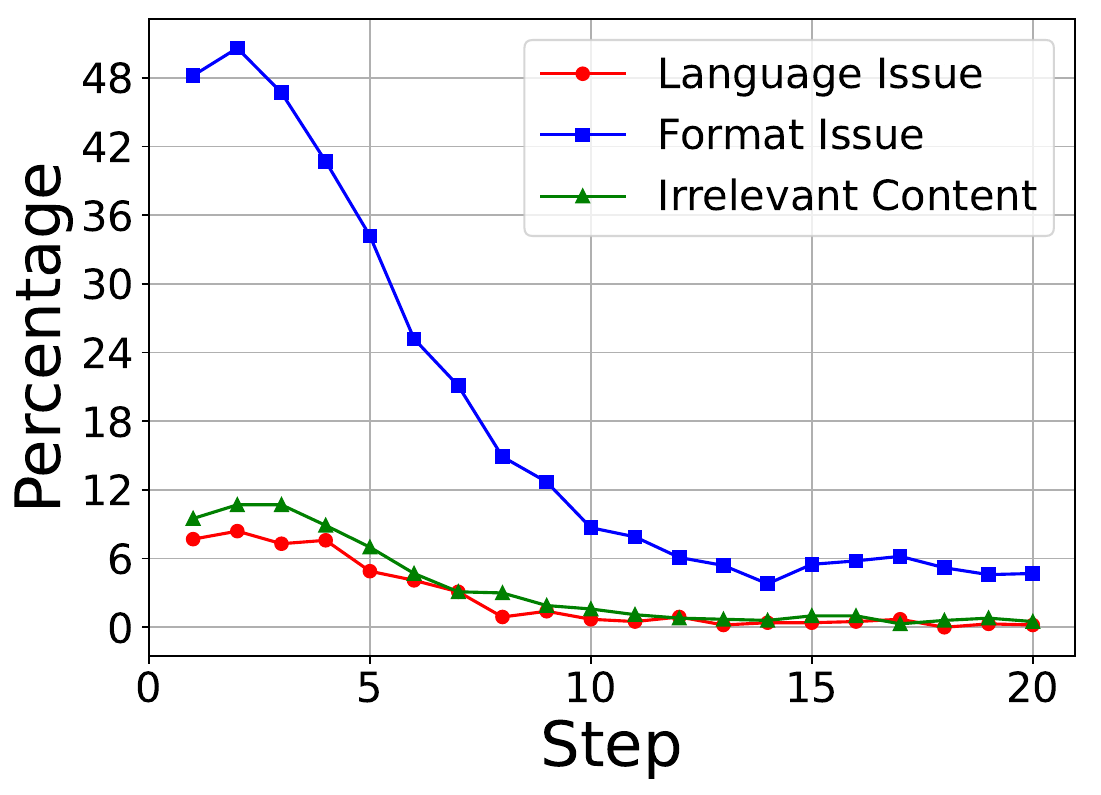}
  }

  \caption{Overview of entropy trends, accuracy evolution, token dynamics, and response issues during GRPO training's rising stage.}
  \label{fig:combined_four_figures}
\end{figure}

\subsubsection{Token-Level Metrics for RL Algorithmic Analysis} 
To enable a deeper analysis of RL algorithms in the RLVR setting, we introduce three fine-grained metrics that quantify token-level algorithmic behavior.  

% \begin{itemize}[leftmargin=0.0em,itemindent=0em]
% \item 
$\bullet$\emph{Entropy.}
Following Section~\ref{sec:ent}, we use token-level entropy $H_t$ (defined in Equation~\ref{eq:token_entropy}) to quantify uncertainty in the policy $\pi_\theta$'s predictions at generation step $t$.
Higher $H_t$ values indicate greater uncertainty in token selection, reflecting exploration potential during generation.

$\bullet$ \emph{Gradient.}
To analyze how tokens drive policy updates, we estimate each token's contribution to policy updates by computing the gradient of the GRPO objective \(J_{\text{GRPO}}(o_t^i)\) with respect to the language model head layer and taking its Frobenius norm as the update magnitude proxy~\citep{wang2025stabilizing}. Formally, the Frobenius norm of the resulting gradient for the $t$-th token is computed as:
\begin{equation}
\small
G_t = \left\| \alpha_t \,\bigl(\bm{e}(o_t) - \bm{\pi}_{\theta}\bigr)\cdot \bm{h}^\top \right\|_F,
\label{eq:grad}
\end{equation}
where \(\alpha_t = \hat{r}_t \cdot \min(\hat{A}_t, \text{clip}(\hat{A}_t, 1-\epsilon, 1+\epsilon))\), \(\bm{e}(o_t)(o_t^i)\in \mathbb{R}^V\) is the one-hot vector for token $o_t$, and \(\bm{\pi}_{\theta}\in \mathbb{R}^V\) is the policy distribution. \(\bm{h} \in \mathbb{R}^d\) is the output of the last transformer layer. The full derivation is in Appendix~\ref{app:gradient_derivation}.

$\bullet$ \emph{Performance Impact.}
To quantitatively assess the impact of tokens on reasoning accuracy, we design a token replacement intervention stragety. For any token \( o_t^i \) within a generated sequence, we substitute it with the highest-probability alternative token under the current policy:
\begin{equation}
\small
\tilde{o}_t^i = \arg\max_{v_k \in V \setminus \{o_t^i\}} \pi_{\theta}(v_k \mid q, o_{<t}).  
\end{equation}
Subsequent $k$ continuations are generated independently from both the original token \( o_t^i \) and the substituted token \( \tilde{o}_t^i \). The divergence in average solution accuracy between these paired continuation paths serves as a metric for the token's influence on downstream reasoning correctness:
\begin{equation}
\small
I_t = \frac{1}{k} \sum_{j=1}^{k} \left( \text{Acc}_j(q, o_{<t}, o_t)\right) -  \frac{1}{k} \sum_{j=1}^{k}\left(\text{Acc}_j(q, o_{<t}, \tilde{o}_t) \right).
\label{eq:token_i}
\end{equation}
Here, Acc(·) is a binary function that returns 1 if the completed sequence leads to a correct solution, and 0 otherwise.
% \end{itemize}

\subsubsection{Stage-level Dynamics}

Prior work~\cite{entropymechanism} identifies two distinct stages in RLVR training dynamics: (1) a rapid \textit{rising stage} with quick performance improvements and decreasing policy entropy, followed by (2) a stable \textit{plateau stage} with marginal gains (Figure~\ref{fig:ent_and_acc}). This bimodal behavior naturally raises the question: what underlying mechanisms drive performance improvements in each stage?

%, erot, whim, Dead, ther, flirt, ĉquery,pornography,
%\_, \textbackslash\}\textbackslash, x, 7, \textbackslash[, ).,sqrt, times,  cdot,**, 9, (\textbackslash, 8, :,5, ), 6,[, 0, 4, \textbackslash), ],
\paragraph{Rising Stage.}

To understand the rapid performance gains in this stage, we analyze the source of entropy reduction and its effects on model behavior. We divide the model responses at each training step into positive and negative sets, and track their entropy dynamics, revealing two main phenomena:

$\bullet$~\textbf{Entropy reduction mainly stems from negative samples.} As shown in Figure~\ref{fig:pos_isu}, negative samples consistently exhibit higher average policy entropy than positive samples. More importantly, their entropy declines at a substantially more rapid rate during the rising stage. Also, tokens that appear exclusively in negative samples experience the fastest decline in entropy.
This suggests that penalizing incorrect reasoning paths plays an important role in the model's initial learning signal, reducing the vast space of potential errors.

$\bullet$~\textbf{Entropy reduction solidifies effective reasoning patterns.} Our analysis of token distributions (Table~\ref{tab:token-table}) reveals that the most significant entropy reductions occur in tokens unrelated to the task objective, while reasoning-critical tokens show increased frequency. Furthermore, we categorize low-quality responses into three types: format violations (unboxed or multiply-boxed answers), irrelevant content (garbled or repetitive text), and language mixing (multilingual responses). For format violations, we count the occurrences of ``\textbackslash\textbackslash boxed\{'' in the response string. To identify irrelevant content, we utilize Qwen2.5-32B-Instruct to determine if the response contains such content. For language mixing, we employ a Regular Expression to check if any token's Unicode encoding falls within the range of Chinese characters. As shown in Figure~\ref{fig:isu}, this entropy shaping leads to a marked decrease in all three key types of defective outputs.

\paragraph{Plateau Stage.}
\begin{figure}[H]
  \centering

  % 左图：概率变化
  \subfigure[Token probability shifts after gradient update.\label{fig:prob_change}]{
    \includegraphics[width=0.43\textwidth]{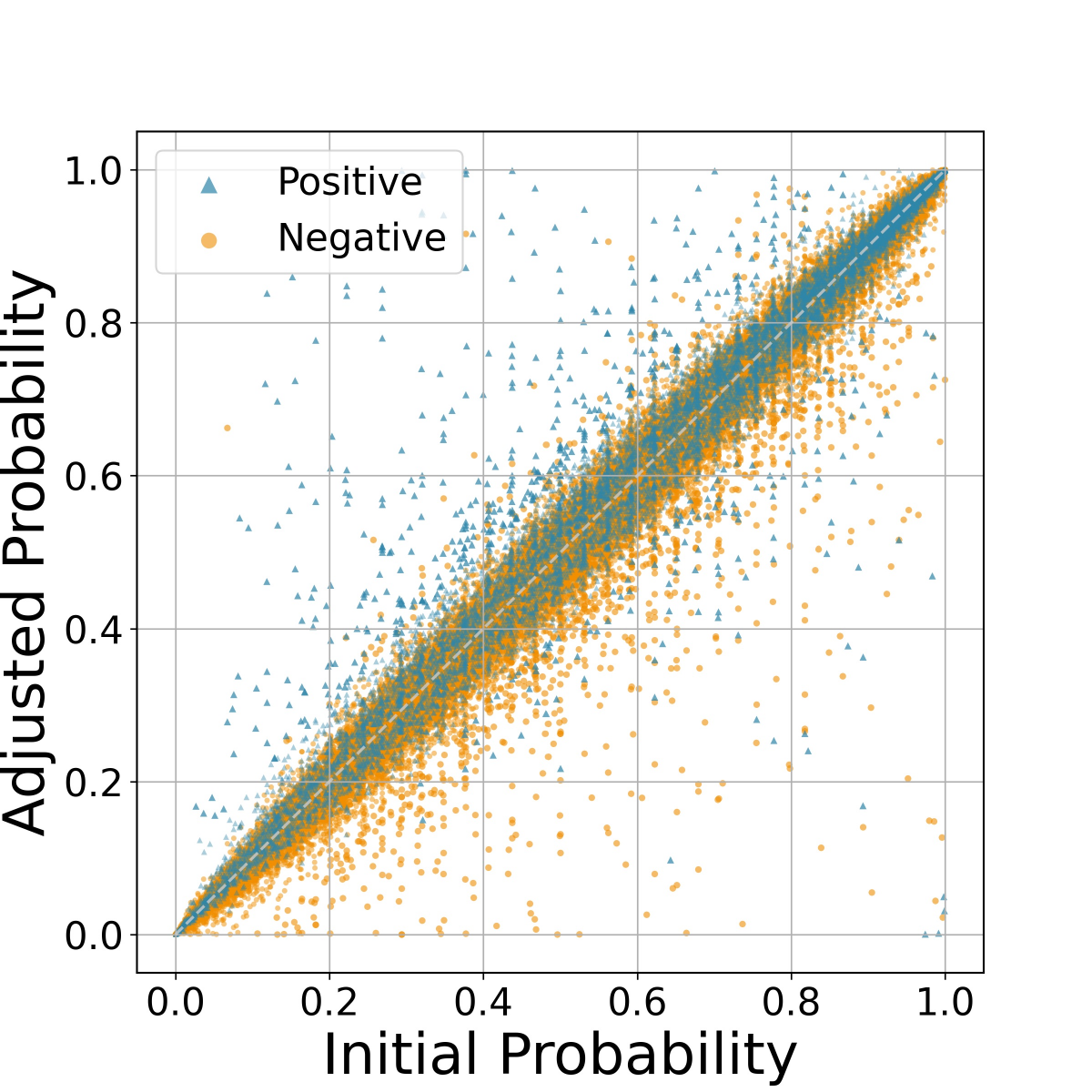}
  }
  \hspace{0.01\textwidth}
  % 右图：熵与梯度分布
  \subfigure[Token entropy and gradient distribution.\label{fig:ent_and_grad}]{
    \includegraphics[width=0.43\textwidth]{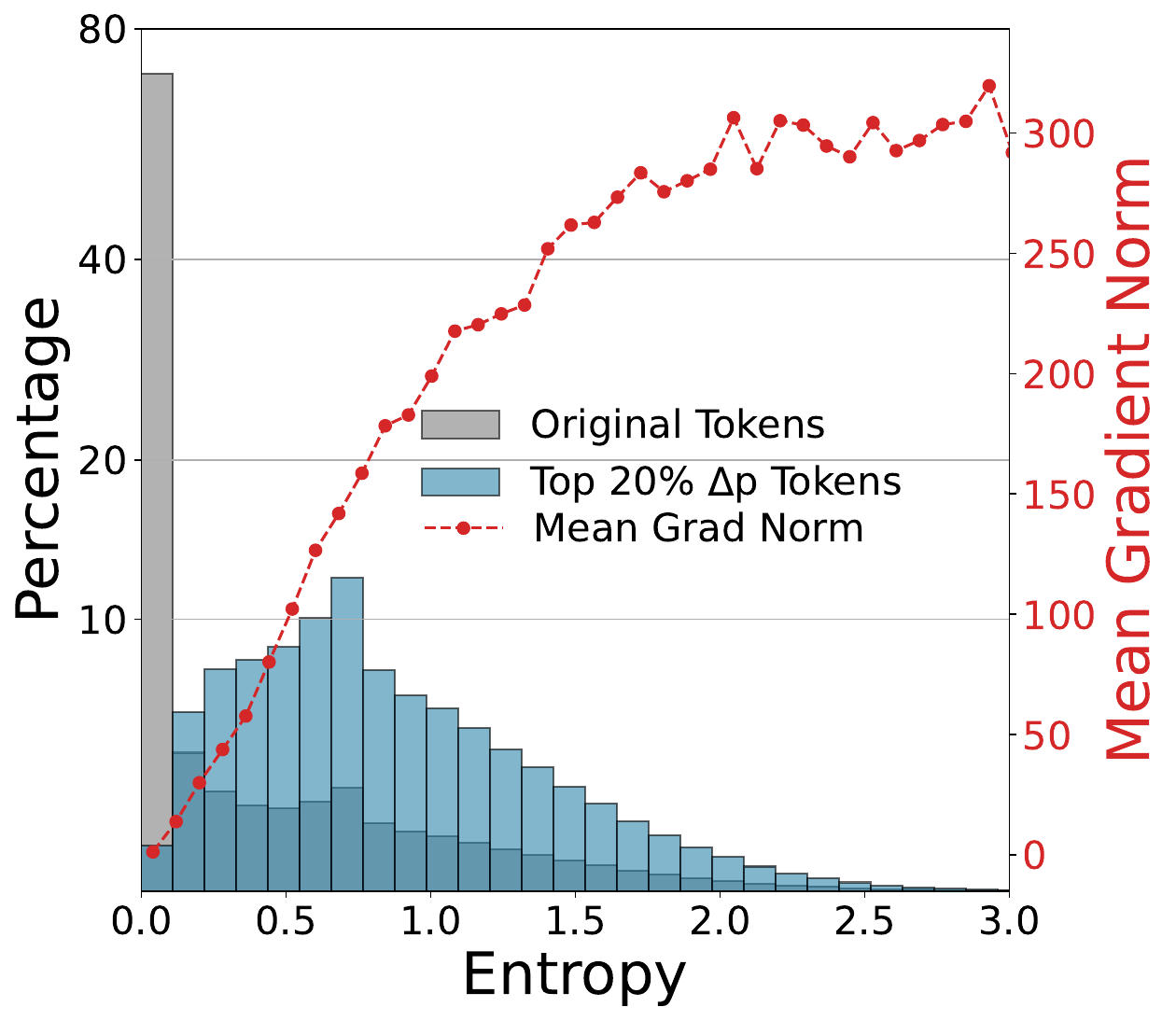}
  }

  \caption{Token-level update patterns.}
  \label{fig:prob_scatter}
\end{figure}

In this stage, as performance gains become incremental and entropy change flattens, 
we conduct a fine-grained investigation into the underlying mechanisms driving continued refinement. Specifically, we examine the distribution of token-level probability updates, analyzing both the magnitude of learning signals received by different tokens and their relationship to entropy dynamics and semantic roles.
%we further investigate fine-grained mechanisms for performance  refinement. Specificially, we analyze the distribution of token probability updates, examining which tokens receive the strongest learning signals and how this relates to their entropy and semantic function.
%单独出来了

$\bullet$~\textbf{Learning concentrates on a small subset of high-entropy, high-gradient tokens.} % 为了分析token的更新特征,我们从平台期随机挑选一些step,并统计该step的训练集中每个token在梯度更新后的概率变化。
Unlike the rising stage, our analysis of token probability updates reveals that most token probabilities remain stable during the plateau stage, with over 99\% of tokens experiencing a probability change of less than 0.06 after parameter updates.
As illustrated in Figure~\ref{fig:prob_change}, learning is instead concentrated on a small fraction of tokens where probabilities in positive samples are reinforced while those in negative samples are suppressed. In Figure~\ref{fig:ent_and_grad}, these impactful updates primarily target high-entropy tokens. These tokens tend to produce larger gradients during backpropagation (Eq.~\ref{eq:grad}). This indicates that progress in this stage is mainly driven by resolving uncertainty at critical ``forks'' in reasoning paths~\cite{beyond82rule}.

$\bullet$~\textbf{Updates are most sensitive for tokens associated with formal reasoning.} To further characterize these critical tokens, we categorize them by their semantic roles and analyze which types experience the largest probability changes: \emph{formal reasoning} tokens enable symbolic manipulation for computation and modeling; \emph{logical structuring} tokens manage the flow of reasoning; \emph{metacognitive} tokens guide the process through self-monitoring; and \emph{semantic support} tokens provide linguistic elements for fluency, coherence, and informativeness.
We provide examples of each token category in Table~\ref{tab:token_examples}. 
Our results show that among the top 20\% of tokens with the greatest probability updates, those associated with formal reasoning (\eg numerals, mathematical symbols) have the highest proportion ($0.039$), followed by metacognitive reasoning tokens ($0.034$), general semantic tokens ($0.033$), and logical structuring tokens ($0.031$). This targeted refinement of critical, uncertain tokens indicates a shift towards mastering the nuanced logic and precise calculations required for advanced reasoning, rather than merely reproducing structural patterns.

\begin{table}[ht]
\centering
\caption{Examples of Token Categories in RLVR.}
\label{tab:token_examples}
\begin{tabular}{l p{0.72\linewidth}}
\toprule
\textbf{Category} & \textbf{Examples} \\
\midrule
\textbf{Formal Reasoning} & Numbers (\eg `1', `3.14'), operators (\eg `+', `*', `='), variables (\eg `x', `y'), and symbols (\eg `$\pi$', `$\sqrt{2}$', `$\sum$'). \\
\\
\textbf{Logical Structuring} & Causal (\eg `therefore', `because'), contrastive (\eg `however', `but'), progressive (\eg `first', `next', `finally'), and parallel (\eg `and', `also'). \\
\\
\textbf{Metacognitive} & Verifying (\eg `Let's check'), revising (\eg `Correction', `Wait'), summarizing (\eg `In summary'), and planning (\eg `First, I will...'). \\
\\
\textbf{Semantic Support} & Grammatical elements (\eg `the', `is', `of'), domain entities (\eg `problem', `solution'), and adjectives (\eg `correct', `final'). \\
\bottomrule
\end{tabular}
\end{table}

\begin{tcolorbox}
[colback=black!5!white,colframe=black!55!white,title=\textbf{\footnotesize \textsc{\ul{Takeaway for Stage-Level Analysis (Section 3.2.2)}}:}]
\begin{itemize}
[leftmargin=1mm]
\setlength\itemsep{0em}
    \item[\ding{224}] {\footnotesize 
    {\fontfamily{phv}\fontsize{8}{9}\selectfont
   During the rising phase, entropy reduction is primarily driven by negative examples, facilitating the emergence of reasoning patterns.}} 
    
    \item[\ding{224}] {\footnotesize 
    {\fontfamily{phv}\fontsize{8}{9}\selectfont
   During the plateau phase, probability shifts are concentrated in a small set of high-entropy, high-gradient tokens associated with formal reasoning.}}
\end{itemize}
\end{tcolorbox}
\subsubsection{Instance-level Efficacy}
As not all samples contribute equally to learning~\cite{chen2024allo}, to understand how instance quality affects optimization, we analyze the role of instance-level PPL, which can be regarded as a measure of the model's uncertainty over a whole sequence. 
Since low-PPL responses are generally more fluent and semantically coherent~\cite{adiwardana2020towards}, we hypothesize that these low-PPL instances are more critical for effective RLVR, which is confirmed by the following three findings from our analysis:
\begin{figure}[!htbp]
  \centering

  % 图1：PPL分布
  \subfigure[Distribution of PPL values for tokens with the top 20\% largest probability shifts in the training set.\label{fig:ppl_distribution}]{
    \includegraphics[width=0.43\textwidth]{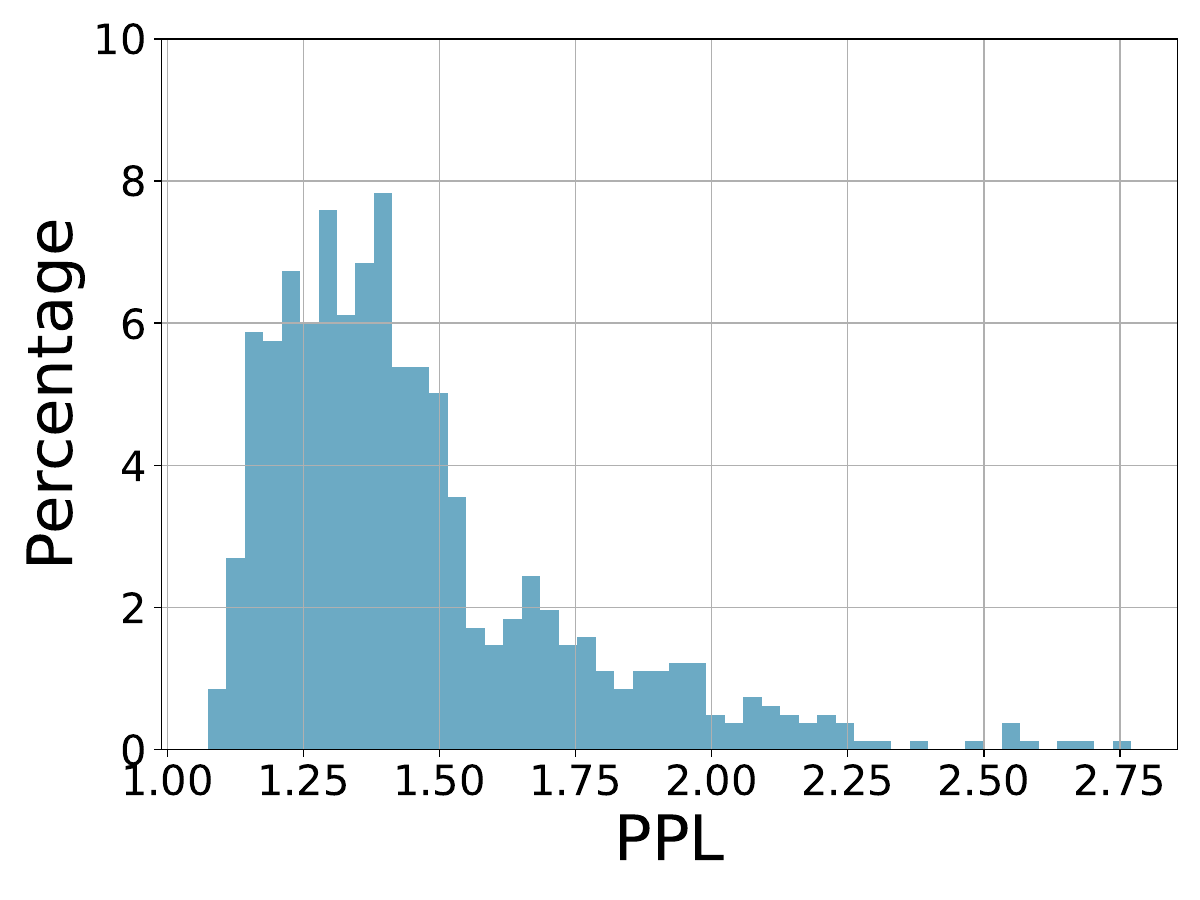}
  }
  \hfill
  % 图2：替换高熵token后的准确率变化
  \subfigure[Accuracy change after replacing high-entropy tokens in samples with high vs. low PPL.\label{fig:ppl_rollout}]{
    \includegraphics[width=0.43\textwidth]{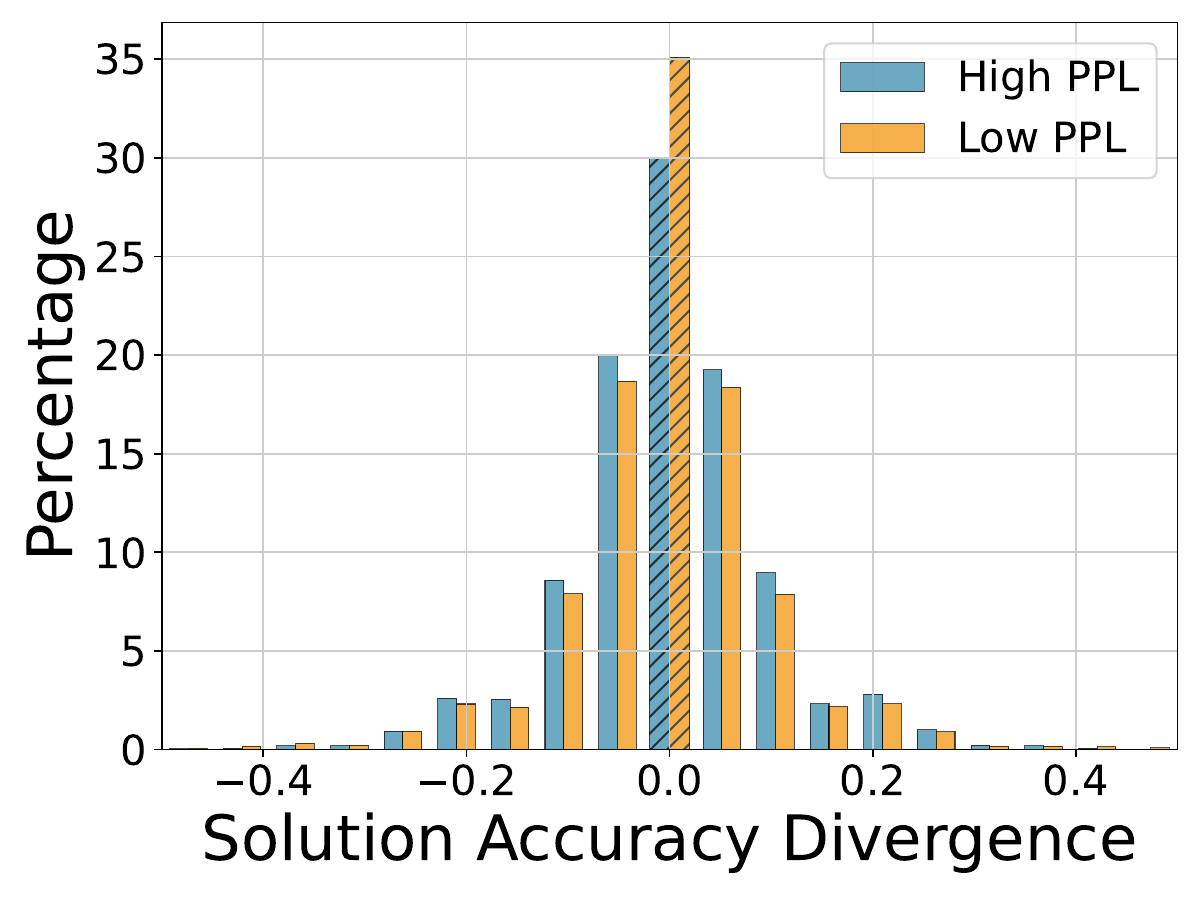}
  }
  \hfill
  % 图3：准确率 vs PPL 奖励
  \subfigure[Model accuracy under advantage schemes favoring high or low PPL examples..\label{fig:abli_ppl}]{
    \includegraphics[width=0.43\textwidth]{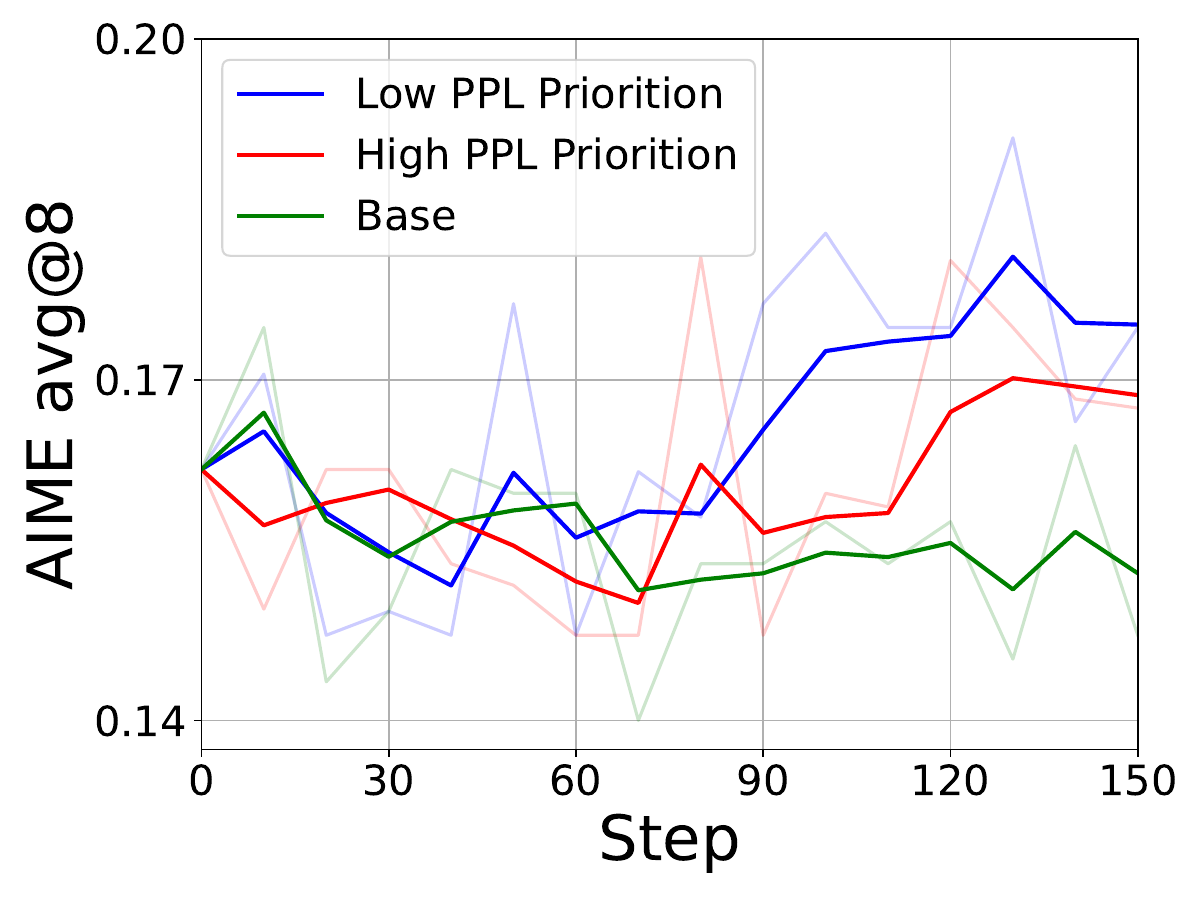}
  }
  \hfill
  % 图4：熵 vs PPL 奖励
  \subfigure[Average token entropy under advantage schemes favoring high or low PPL examples.\label{fig:high-low-ppl-entropy}]{
    \includegraphics[width=0.43\textwidth]{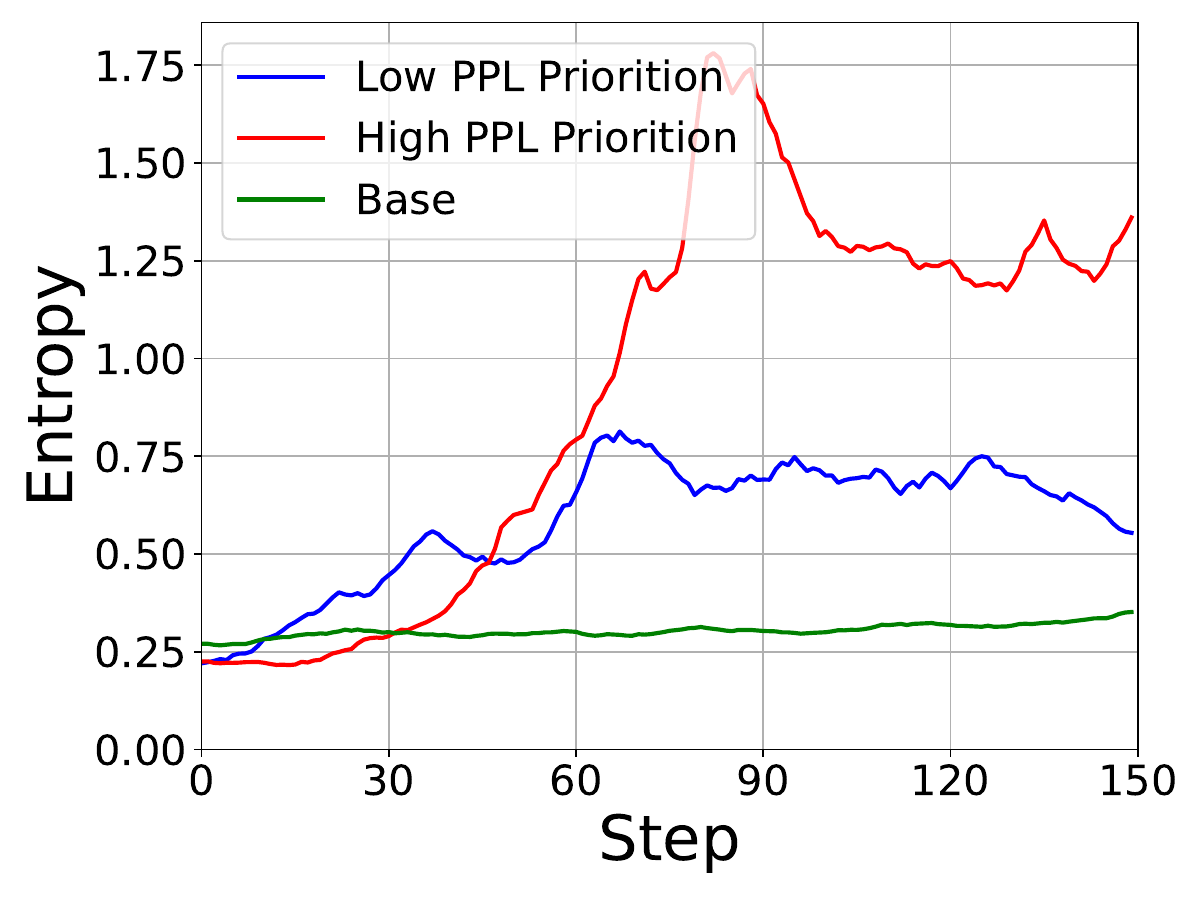}
  }

  \caption{Analysis of token behavior and model performance under different manipulations related to PPL.}
  \label{fig:ppl_analysis_combined}
\end{figure}

$\bullet$ \textbf{Learning signals are concentrated in low-PPL samples.}
To explore where learning occurs most actively, we analyze the magnitude of token probability changes during RLVR updates. As shown in Figure~\ref{fig:ppl_distribution}, we observe a clear concentration of high-magnitude probability updates in the low-PPL region, indicating that the model's learning is more active within these generations.

$\bullet$ \textbf{Low-PPL instances represent more robust reasoning paths.} To understand the differences between samples, we apply token-level intervention analysis (Eq.~\ref{eq:token_i}) to instances sampled from both low-PPL (bottom 20\%) and high-PPL (top 20\%) groups. The results in Figure~\ref{fig:ppl_rollout} show that replacing tokens in low-PPL responses leads to smaller changes in the final solution's accuracy compared to the same intervention in high-PPL responses, indicating that the model exhibits more robust and stable reasoning in low-PPL instances.

$\bullet$~\textbf{Prioritizing low-PPL instances enhances RLVR effectiveness.} To verify the importance of low-PPL instances, we conduct the experiment by dynamically re-weighting token advantages based on PPL. First, we compute a standardized log-PPL weight for each response $o^i$:
\begin{equation}
\small
w_{\text{ppl}}(o^i) = \frac{\ln \mathrm{PPL}(o^i) - \mu}{\sigma}.
\label{eq:ppl_w}
\end{equation}

Here \( \mu \) and \( \sigma \) are the mean and standard deviation of the log-PPL values across the sampled responses for the same query \( q \), and $\alpha$ is a hyperparameter. We then compare two opposing strategies: one that adjusting the advantage with a factor of $(1 - \alpha \cdot w_{\text{ppl}}(o^i))$ of sampled instances, and another that using the factor of $(1 + \alpha \cdot w_{\text{ppl}}(o^i))$. As shown in Fig.~\ref{fig:abli_ppl}, the former one results in superior performance gains.
In contrast, focusing on high-PPL samples leads to much higher policy entropy, as shown in Figure~\ref{fig:high-low-ppl-entropy}. Further analysis of the model's generated responses on the test set reveals that this approach degrades response quality, with the frequency of responses containing quality issues rising to approximately 7\%, compared to about 3\% for the low-PPL strategy. This confirms that focusing RL updates on low-PPL samples is a more effective optimization strategy.
\begin{tcolorbox}
[colback=black!5!white,colframe=black!55!white,title=\textbf{\footnotesize \textsc{\ul{Takeaway for Instance-Level Analysis (Section 3.2.3)}}:}]
\begin{itemize}
[leftmargin=1mm]
\setlength\itemsep{0em}
    \item[\ding{224}] {\footnotesize 
    {\fontfamily{phv}\fontsize{8}{9}\selectfont
   Low PPL samples are crucial for model self-improving, as they exhibit larger token probability shifts, demonstrate more robust and stable behavior, and offer higher optimization efficiency.}} 

\end{itemize}
\end{tcolorbox}
\subsubsection{Token-level Significance}

To understand how a token's effect on learning varies throughout a sequence, we analyze the interplay between token position, entropy, and optimization impact. We investigate the distribution of token entropy and importance across different positions, finding that although entropy is high at both the beginning and end of sequences, the tokens toward the end are more critical for effective RL.
\begin{figure}[!htbp]
  \centering

  % 图1：位置熵分布
  \subfigure[Average token entropy by position across the training set, computed over 1k steps.\label{fig:pos_ent}]{
    \includegraphics[width=0.43\textwidth]{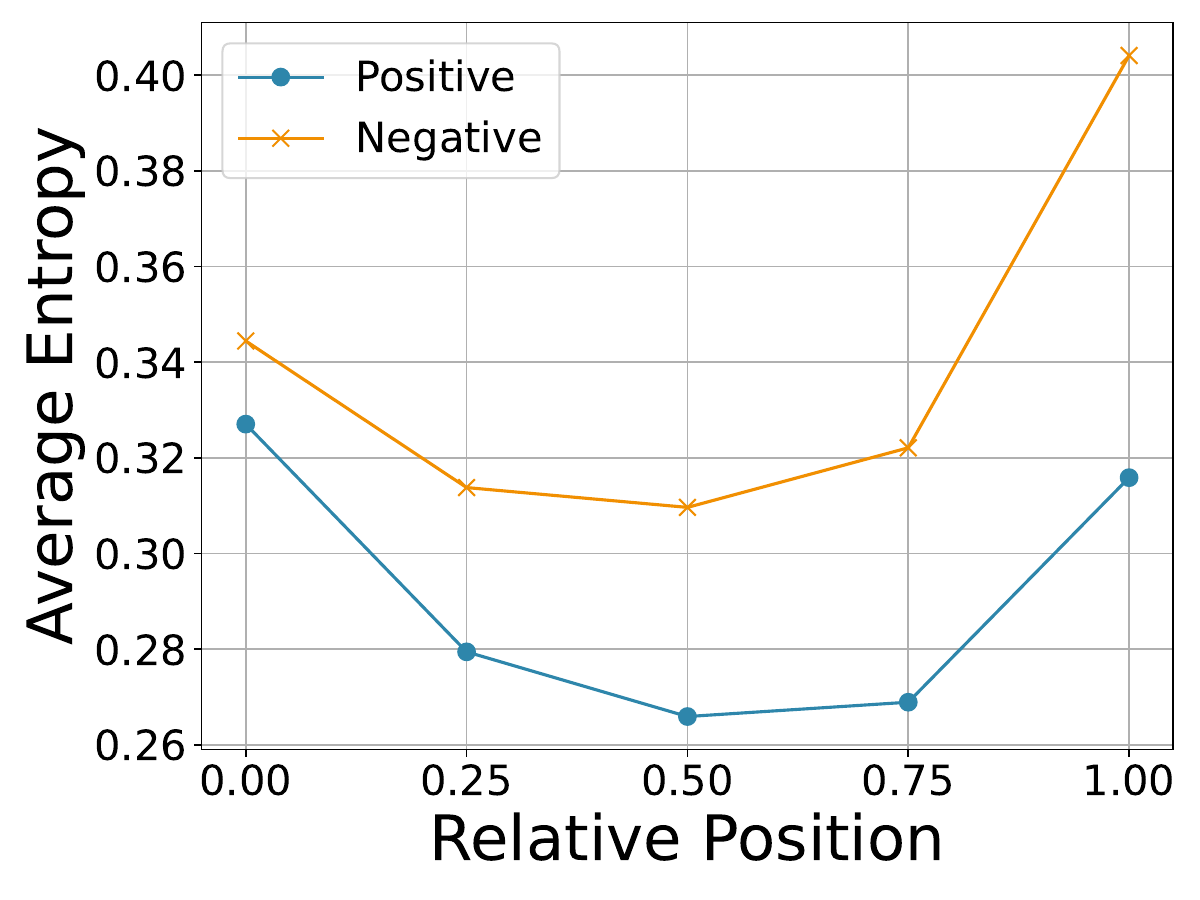}
  }
  \hfill
  % 图2：替换高熵 token 导致的准确率变化
  \subfigure[Accuracy change from replacing high-entropy tokens at top and bottom 20\% positions.\label{fig:loc_rollout}]{
    \includegraphics[width=0.43\textwidth]{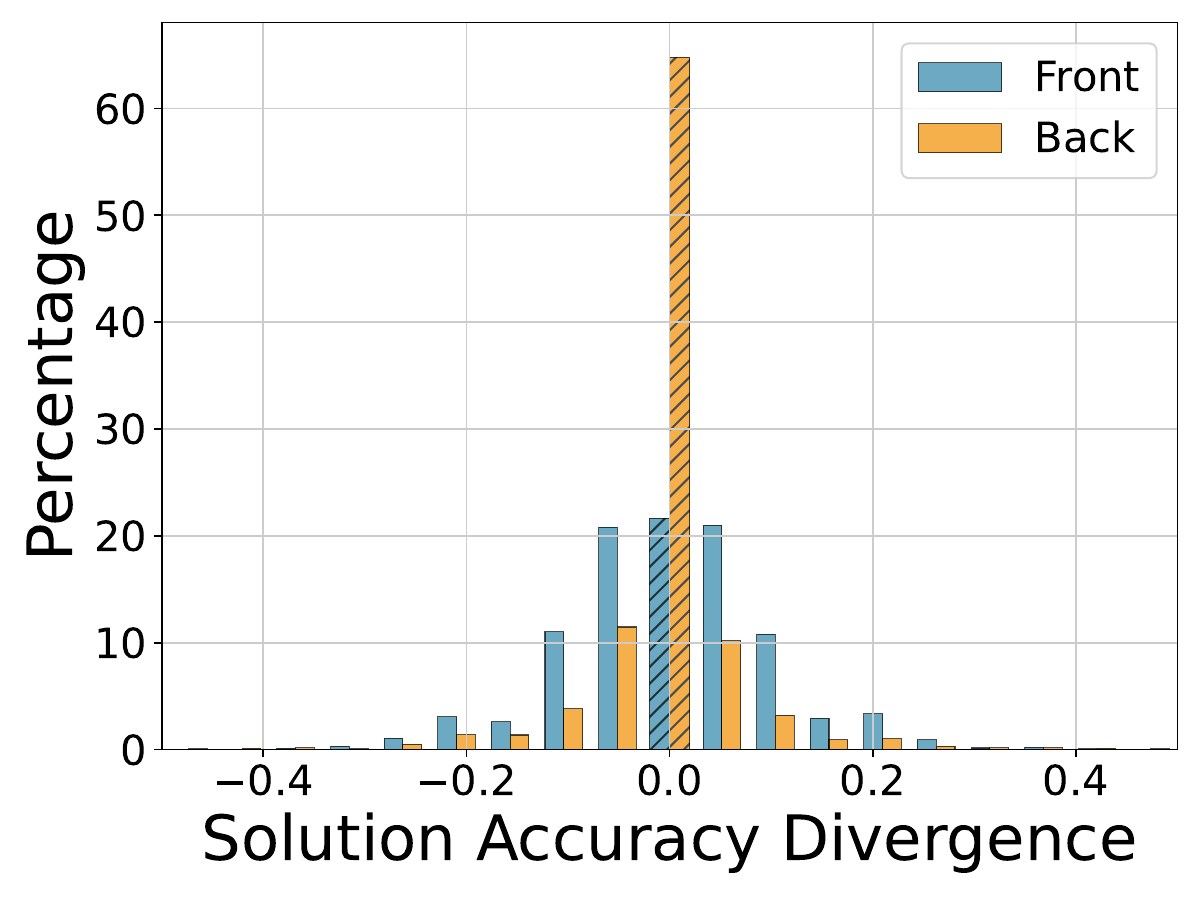}
  }
  \hfill
  % 图3：位置奖励对准确率的影响
  \subfigure[Model accuracy under positional advantage schemes with fixed $\gamma = 1.0$..\label{fig:three_curves2}]{
    \includegraphics[width=0.43\textwidth]{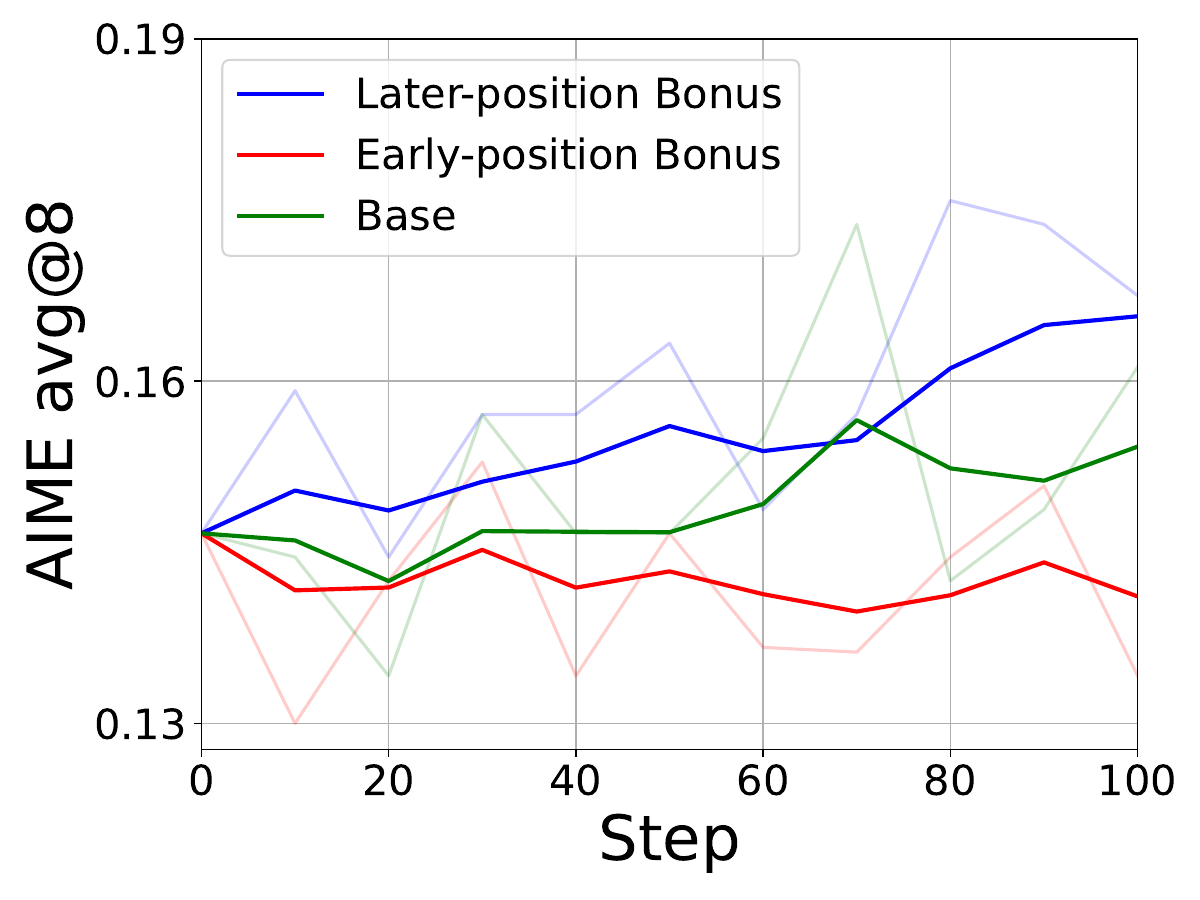}
  }
  \hfill
  % 图4：位置奖励对熵的影响
  \subfigure[Average token entropy under positional advantage schemes with fixed $\gamma = 1.0$.\label{fig:loc_entropy}]{
    \includegraphics[width=0.43\textwidth]{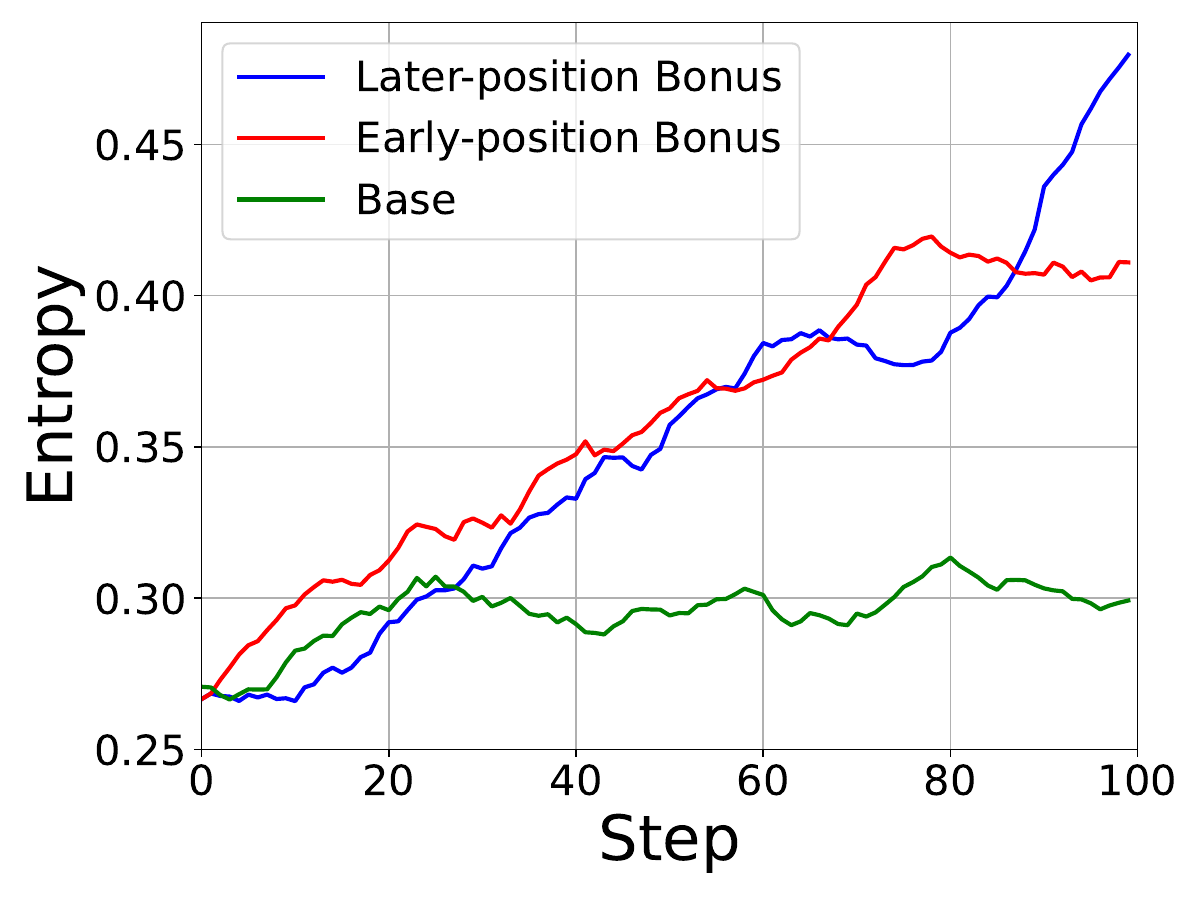}
  }

  \caption{Token position analysis and model performance under position-based reward schemes.}
  \label{fig:position_analysis}
\end{figure}

$\bullet$ \textbf{Token entropy follows a U-shaped distribution, with higher values at the start and end of sequences.} 
As illustrated in Fig~\ref{fig:pos_ent}, we observe that higher entropy concentrates at the beginning and end of a response. High entropy at the beginning reflects a broad exploration space where the model considers multiple initial approaches. In contrast, high entropy near the end of a sequence indicates uncertainty in the final decision-making process, which is directly linked to the task objective. As noted in prior work~\cite{prabhudesai2025maximizing}, there is a high correlation between model confidence in the last few tokens and overall accuracy.

$\bullet$ \textbf{Initial high-entropy tokens govern outcomes; terminal high-entropy tokens reflect reasoning  uncertainty.}
We use token-level intervention analysis in Eq.~\ref{eq:token_i} and reveal the distinct functional roles of these two high-entropy regions.
As Figure~\ref{fig:loc_rollout} illustrates, replacing early-position tokens significantly alters the final solution's accuracy. This highlights the inherent uncertainty in the initial language space, which broadens the exploration scope and results in higher entropy. Conversely, while late-position tokens also exhibit high entropy, their minimal impact on accuracy suggests a more constrained semantic space. Interestingly, the entropy of late-position tokens in negative examples is higher than in positive ones. This subtly indicates that the model might, in the later stages of inference for incorrect solutions, implicitly detect its errors, leading to greater confusion and, consequently, elevate entropy.

$\bullet$~\textbf{Optimizing tokens in later positions provides a more efficient learning signal.} To verify this, we conduct a comparative experiment by applying a positional bonus to the token advantages, defined as follows:
\begin{equation}
\small
b^i_t = \gamma \cdot \sigma(d \cdot r^i_t).
\label{eq:loc-bonus}
\end{equation}
where $\gamma$ is a hyperparameter, \( \sigma \) is the sigmoid function, $r^i_t$ represents the token's relative position, and the direction parameter $d$ determines the focus of the bonus. Setting $d=1$ rewards tokens appearing later in the sequence, while setting $d=-1$ rewards tokens appearing earlier. For positive samples, this bonus is added to the original advantage to increase the reward, while for negative samples, it is subtracted to amplify the penalty. Our experiment results in Fig.~\ref{fig:three_curves2} shows that reinforcing tokens later in the sequence yields superior performance compared to both baselines with no positional bonus and the strategy that gives bonuses to early tokens.
While applying the positional bonus in either direction increases policy entropy (Figure~\ref{fig:loc_entropy}), further analysis of the generated responses reveals that rewarding early positions leads to shorter average response lengths (904 tokens) compared to rewarding later positions (1146 tokens). This suggests that optimizing the latter parts of reasoning can extend the model's reasoning time~\citep{DeepSeek-R1}, thereby improving accuracy.
\begin{tcolorbox}
[colback=black!5!white,colframe=black!55!white,title=\textbf{\footnotesize \textsc{\ul{Takeaway for Token-Level Analysis (Section 3.2.4)}}:}]
\begin{itemize}
[leftmargin=1mm]
\setlength\itemsep{0em}
    \item[\ding{224}] {\footnotesize 
    {\fontfamily{phv}\fontsize{8}{9}\selectfont
   Positions towards the end are more crucial for model learning, as they have higher entropy, more stable semantics, and better optimization efficiency.}} 
\end{itemize}
\end{tcolorbox}
\section{Exploration-Enhanced RL Approaches} 
\label{sec:performance improvement}
In this section, we discuss how to enhance model performance using exploration-enhanced reinforcement learning (RL) approaches. We start by introducing the RLVR baseline—the GRPO method—followed by a literature survey of related methods. Next, we present experimental results and discuss strategies to enhance exploration capabilities based on the Pass@k metric. Finally, we introduce several simple methods to improve RL performance, drawing on the findings from Section~\ref{sec:inter}.

\subsection{The RLVR Baseline -- GRPO}%  (cdx)

GRPO~\citep{grpo} is a representative RL algorithm for LLMs. It optimizes a policy \(\pi_\theta\) to maximize the expected reward over sampled reasoning trajectories. Given an old policy \(\pi_{\theta_{\text{old}}}\) and the current policy \(\pi_\theta\), GRPO maximizes the following objective:
\begin{align}
\mathcal{J}(\theta) =
\mathbb{E}_{q \sim \mathcal{D},\; o \sim \pi_{\theta_{\text{old}}}} \Bigg[&
\sum_{t=1}^{|o|}
\min\Big( r_t \hat{A}_t,\; 
\text{clip}(r_t,\; 1\!-\!\epsilon,\; 1\!+\!\epsilon) \hat{A}_t \Big) 
\notag \\
&- \beta \cdot \mathrm{KL}\big[\pi_\theta(\cdot \mid q, o_{<t}) \,\|\, \pi_{\text{ref}}(\cdot \mid q, o_{<t})\big]
\Bigg],
\end{align}
where \(q\) and \(o\) denote the input prompt and the sampled response respectively, drawn from the prompt dataset \(\mathcal{D}\) and the old policy \(\pi_{\theta_{\mathrm{old}}}\). The importance sampling ratio \(r_t\) is
\(
r_t = \frac{\pi_\theta(o_t | q, o_{<t})}{\pi_{\theta_{\mathrm{old}}}(o_t | q, o_{<t})},
\) and \(\hat{A}_t\) denotes the token-level advantage estimate. The hyperparameter \(\epsilon\) controls the clipping range, and \(\beta\) weights the KL regularization term against a fixed reference policy \(\pi_{\mathrm{ref}}\).

To compute \(\hat{A}_t\), GRPO applies a group-relative normalization scheme. In outcome-supervised settings, for each prompt \(q\), it samples \(G\) responses \(\{o^1, o^2, \dots, o^G\}\) using the old policy \(\pi_{\mathrm{old}}\), and assigns each response a binary reward \(R^i\): 1 if correct and 0 otherwise. Since the reward is uniformly broadcast over all tokens, the token-level advantage for the \(t\)-th token in the \(i\)-th response \(o^i\) is

\begin{equation}
\hat{A}_t^i = \frac{R^i - \mathrm{mean}(\{R^j\}_{j=1}^G)}{\mathrm{std}(\{R^j\}_{j=1}^G)}.
\end{equation}

\subsection{The Exploration-Exploitation Trade-off in RLVR} 
The core of RLVR lies in balancing exploration and exploitation—a classic challenge in reinforcement learning algorithms. Initially, the model has limited knowledge of the target task and can only make meaningful attempts based on prior knowledge (\ie knowledge acquired from training data). The environment or verifier then provides feedback to guide the model's learning~\cite{shao2024deepseekmath,schulman2017proximal}. Unlike SFT, however, all training data involved in RLVR—whether correct or incorrect solutions—are generated by the model itself. A key distinction between RL and SFT is thus: RL reinforces correct behaviors while discouraging incorrect ones, whereas SFT merely imitates correct demonstrations~\cite{chu2025sft}. This makes RL focus on learning from self-generated behaviors; over time, this can lead to overconfidence in intermediate actions, which in turn leads to entropy collapse and performance saturation~\cite{entropymechanism}.

Once the model has acquired certain skills through self-generated data, developing new skills requires exploratory behaviors to attempt novel approaches. This dynamic embodies the fundamental exploration-exploitation trade-off. For strong performance in RLVR, two critical factors emerge: first, the model must be capable of diverse, meaningful exploration, with a sufficiently large exploration space to sample correct solutions; second, the model should be trained efficiently to identify correct solutions with minimal effort. 
These considerations give rise to two key research questions: \emph{how to enhance LLMs' exploration capabilities} and \emph{how to translate these exploration abilities into performance gains?}

In what follows, we first review the previous work that attempts to enhance the RLVR method in Section~\ref{sec:related_work_enhance_explora}. Then we  conduct the empirical experiments to enhance the  model capabilities in Section~\ref{sec:empirical_study_enhance_ability}.

\subsection{Reviewing Existing Studies on Enhancing Exploration}
\label{sec:related_work_enhance_explora}

To improve the training performance of RLVR, recent studies have proposed methods to preserve or enhance the model’s exploration capacity during training. These approaches aim to mitigate distributional narrowing—evidenced by declining policy entropy or Pass@k metrics—ensuring the model retains sufficient reasoning diversity to discover and refine correct solutions. 
Concretely, these methods intervene in various components of the RL pipeline, including advantage shaping, token-level gradient modulation, KL-based regularization, and external tool integration. Below, we categorize and summarize these exploration-enhancing techniques across four primary areas, as detailed in Table~\ref{tab:review_extend_exploration}.

\begin{table}[!htb]
\centering
\caption{Exploration enhancement methods in different categories. We also report the metrics for measuring the exploration capacities.}
\label{tab:review_extend_exploration}
\vspace{3pt}
\begin{tabular}{lll}
\toprule
\textbf{Category} & \textbf{Method} & \textbf{Metric} \\
\midrule
\multirow{3}{*}{Advantage Refinement} 
    & Entropy-based Advantage Shaping~\citep{reason_explore} & Pass@k \\
    & Negative Sample Reinforcement~\citep{nsr} & Pass@k \\
    & Pass@k Training~\citep{passk_training} & Pass@k \\
\midrule
\multirow{3}{*}{Token/Gradient Selection} 
    & Clip-Higher~\citep{dapo} & Entropy \\
    & Covariance-aware Gradient Detach~\citep{entropymechanism} & Entropy \\
    & Forking Tokens~\citep{beyond82rule} & Entropy \\
\midrule
\multirow{2}{*}{KL Regularization} 
    & KL Reference Reset~\citep{prorl} & Pass@k \\
    & Covariance-aware KL Penalty~\citep{entropymechanism} & Entropy \\
    \midrule
    \multirow{2}{*}{Tool Augmentation} 
    & Progressive Tool Integration~\citep{bai2025towards} & Pass@k \\
    & Entropy-based Adaptive Rollout~\citep{dong2025agentic} & Entropy \\
\bottomrule
\end{tabular}
\end{table}

\paragraph{Advantage Refinement.} 
Modifying the advantage term during policy optimization is a straightforward yet effective approach to enhance LLMs’ exploration ability. Entropy-based Advantage Shaping~\cite{reason_explore} introduces a clipped, gradient-detached entropy bonus into the advantage function, which boosts learning signals for high-entropy (\ie exploratory) tokens while maintaining optimization stability. Negative Sample Reinforcement~\cite{nsr} decomposes the RL objective into positive and negative sample reinforcement, explicitly penalizing incorrect completions to suppress wrong reasoning paths and preserve exploration space. Pass@k Training~\cite{passk_training} extends the advantage estimation to reward all correct completions within top-$k$ outputs, ensuring that exploration of multiple correct solutions is retained before fine-tuning the model toward precise Pass@1 performance. These methods share a common goal: enhancing exploration by ensuring that the advantage term continues to reward diverse reasoning paths throughout the RL training process.

\paragraph{Token/Gradient Selection.}  
Exploration can be improved by selectively controlling token-level gradient updates. The {Clip-Higher} method~\cite{dapo} relaxes the upper clipping bound on the importance sampling ratio in PPO/GRPO, allowing low-probability tokens to increase their likelihood more freely and thus preventing early entropy collapse. {Covariance-Aware Gradient Detach}~\cite{entropymechanism} identifies tokens with unusually high covariance between their log-probabilities and advantages, clipping a small fraction of these tokens out from the gradient computation to avoid overfitting and maintain entropy. {Forking Tokens Selection}~\cite{beyond82rule} focuses optimization on a top fraction of high-entropy tokens within each batch, discarding gradients from low-entropy tokens to concentrate learning on key decision points that promote diverse reasoning paths. All these methods aim to preserve tokens with high exploratory value during training to sustain policy diversity and effective exploration.

\paragraph{KL Regularization.}  
KL-based methods stabilize exploration by penalizing divergence between the current policy and a reference policy, preventing excessive policy drift and improving training stability. ProRL~\citep{prorl} applies a KL penalty throughout training but periodically resets the reference policy to the current policy and reinitializes optimizer states. This reset prevents the KL term from dominating the loss and freezing updates, enabling longer and more stable training while maintaining exploration. {Covariance-Aware KL Penalty}~\citep{entropymechanism} targets tokens with the highest covariance between their log-probabilities and advantages, applying intensified KL penalties only to a small subset of these tokens. By selectively regularizing these influential tokens, KL-Cov prevents premature entropy collapse without overly constraining the entire policy. Together, these techniques balance global and token-level control over policy shifts to sustain exploration and avoid overfitting during RL training.

\paragraph{Tool Augmentation.} 
Integrating external tools into LLM reasoning processes effectively enhances exploration capabilities. ETIR~\cite{bai2025towards} demonstrates that code-integrated reasoning—by executing generated code with external interpreters—significantly expands the model’s capability boundaries and improves exploration, as it enables the model to access new reasoning pathways beyond text generation alone. The close coupling of code generation and execution encourages the model to explore diverse solutions that would be unreachable without tool interaction. ARPO~\cite{dong2025agentic} leverages an entropy-based adaptive rollout strategy to dynamically allocate exploration resources during tool interactions. By monitoring token-level entropy changes after each tool call, ARPO selectively branches new reasoning paths where uncertainty increases, thereby focusing exploration on promising but uncertain regions. This entropy-guided sampling enhances the model’s ability to discover effective multi-step tool-use strategies and improves overall exploration efficiency. 

The methods reviewed above focus on mitigating the exploration space collapse that inherently occurs during RLVR training. By modifying advantage computation, selectively regulating token updates,  applying targeted regularization, these approaches help maintain a sufficiently wide exploration space, which is essential for enabling diverse reasoning strategies. Preserving this exploration capacity sets the foundation for RL to effectively search, discover, and eventually refine correct solutions, ensuring that exploration potential can be effectively translated into performance gains on Pass@1 performance.

\subsection{Empirical Experiments on Enhancing Model Capabilities}
\label{sec:empirical_study_enhance_ability}

In the preceding part, we provided a brief overview of existing progress in enhancing the effectiveness of reinforcement learning training. Here, we present empirical experiments we conducted to examine the efficacy of various strategies for improving model performance. Specifically, we consider two lines of experiments: the first focuses on maintaining the exploratory capabilities of LLMs during the training process, while the second explores more effective training approaches based on the findings in Section~\ref{sec:inter}.

\subsubsection{Maintaining Exploration Capabilities: Retaining Pass@k}

In Section~\ref{sec:explo}, we have compared the effects of SFT and RL noting a key distinction: SFT utilizes external demonstration data, whereas RL relies on self-generated data. When trained on self-generated data, the uncertainty of LLMs is gradually suppressed, leading to a significant reduction in their exploratory capabilities. Accordingly, this section focuses on strategies to preserve such exploratory capacities, with Pass@k serving as the core evaluation metric. 

\paragraph{Methods.}
Considering the high computational cost of RL, we conduct our experiments using Rejection-sampling Fine-Tuning (RFT). RFT iteratively refines a model by sampling multiple responses from the model itself, evaluating those responses, and then fine-tuning the model on a selected subset of these samples. To maintain and enhance the model's exploration capabilities during this process, we propose three data selection strategies for each RFT iteration, applied on top of standard rule-based filtering:

$\bullet$ \emph{Incorporating Noisy Data.} To prevent the model from becoming overly deterministic and to encourage a broader exploration space, we integrate a small proportion of ``noisy'' or incorrect samples into the training data. Specifically, after initial rule-based filtering, we include 5\% of negative samples from the generated rollouts into the training batch.

$\bullet$ \emph{Selecting High-Entropy Data.} High entropy in a generated response indicates greater uncertainty and diversity in the model's token distribution, reflecting a more exploratory behavior. For each candidate response obtained through rejection sampling, we calculate the average token-level entropy across all its tokens. We then prioritize and select a batch of samples for fine-tuning that exhibit the highest average entropy.

$\bullet$ \emph{Selecting High-Rollout Branching Factor (RBF) Data.} Building on the concept of token-level entropy, the rollout branching factor (as defined in Section~\ref{sec:ent}) quantifies the diversity of plausible next tokens at each generation step. We compute the average rollout branching factor for each response and select samples with the highest average values for fine-tuning.

\paragraph{Experimental Setup.}
We conduct experiments using the Qwen2.5-32B model and the STILL-3 dataset. Our RFT process involves multiple iterations, with each iteration utilizing 1.1k selected data points for fine-tuning. The fine-tuning parameters are set as follows: a maximum context length of $20000$ tokens, a batch size of $32$, and a learning rate of $1 \times 10^{-5}$. 
Before applying the specific data selection strategies, we employ a rigorous rule-based filtering process to ensure the quality and relevance of the self-generated responses. This filtering is primarily applied to all samples from the rollouts. The filtering criteria includes:

$\bullet$ \emph{Answer Box Presence:} Only responses containing a properly formatted answer within ``\textbackslash boxed{}'' are considered. Responses without this specific format are discarded.

$\bullet$ \emph{Numerical Content:} For mathematical reasoning tasks, responses are required to contain at least one digit to ensure relevance to the problem type.

$\bullet$ \emph{Post-Answer Content Truncation:} Considering that the base model might produce redundant output after generating the ``\textbackslash boxed{}'' answer, we process all responses by removing any content that followed the first ``\textbackslash boxed{}'' marker.

$\bullet$ \emph{Redundancy Check:} A 3-gram redundancy metric is calculated for each response. Responses with a 3-gram redundancy exceeding 0.31 (indicating high repetition or garbled text) are excluded to maintain content quality.

These rule-based filters ensure that only well-formed, relevant, and non-repetitive incorrect responses are considered for further processing and selection by our proposed strategies.

For evaluation, we assess the models on three benchmarks: AIME24, AMC23, and MINERVA. During evaluation, we set top-p$=0.95$ and temperature $0.6$, sampling $128$ times for each problem on AIME24 and $64$ times for the other two benchmarks to estimate Pass@1 and Pass@k.

% % The experimental results are summarized in Table~\ref{tab:rft_results}.
% \begin{table}[htbp]
% \centering
% \caption{Comparison of Pass@1 and Pass@k performance of Qwen2.5-32B with different RFT data selection strategies on AIME24, AMC23, and MINERVA benchmarks.}
% \label{tab:rft_results}
% \begin{tabular}{lcccccc}
% \toprule
% \multirow{2}{*}{\textbf{Method}} & \multicolumn{2}{c}{\textbf{AIME24}} & \multicolumn{2}{c}{\textbf{AMC23}} & \multicolumn{2}{c}{\textbf{MINERVA}} \\
% \cmidrule(lr){2-3} \cmidrule(lr){4-5} \cmidrule(lr){6-7}
% & Pass@1 & Pass@k & Pass@1 & Pass@k & Pass@1 & Pass@k \\
% \midrule
% Qwen2.5-32B (Base) & \\
% Qwen-2.5-32B-SimpleRL-Zoo & \\
% RFT (Rule-filtered only) & \\
% RFT + Noisy Data (5\%) & \\
% RFT + High-Entropy Data & \\
% RFT + High-RBF Data & \\
% \bottomrule
% \end{tabular}
% \end{table}

\begin{table}[htbp]
\centering
\small
% \setlength{\tabcolsep}{3pt}
% \caption{Comparison of Pass@1 and Pass@k performance of Qwen2.5-32B with different RFT data selection strategies on AIME24, AMC23, and MINERVA benchmarks.}
\caption{Comparison of Pass@1 and Pass@k performance of Qwen2.5-32B with different RFT data selection strategies on AIME24, AMC23, and MINERVA benchmarks. ``Intermediate Methods'' are experiments run for fewer iterations, while the ``Final Method'' is our best combined approach trained for the full duration.   ``RBF'' refers to the rollout branching factor.}
\label{tab:rft_results}
\begin{tabular}{lcccccc}
\toprule
\multirow{2}{*}{\textbf{Method}} & \multicolumn{2}{c}{\textbf{AIME24}} & \multicolumn{2}{c}{\textbf{AMC23}} & \multicolumn{2}{c}{\textbf{MINERVA}} \\
\cmidrule(lr){2-3} \cmidrule(lr){4-5} \cmidrule(lr){6-7}
& Pass@1 & Pass@k & Pass@1 & Pass@k & Pass@1 & Pass@k \\
\midrule
\multicolumn{7}{l}{\textit{Baselines}} \\
Qwen2.5-32B (Base) & 10.50 & 60.00 & 46.60 & 95.00 & 27.50 & 63.52 \\
Qwen-2.5-32B-SimpleRL-Zoo & 27.20 & 65.75 & 67.50 & 95.00 & 36.38 & 69.85 \\
\midrule
\multicolumn{7}{l}{\textit{Intermediate Methods (fewer iterations)}} \\
RFT (Rule-filtered only) & 20.29 & 58.76 & 56.17 & 92.50 & / & / \\
RFT + Noisy Data (5\%) & 18.62 & 63.02 & 57.24 & 92.50 & / & / \\
RFT + High-Entropy Data & 16.43 & 68.30 & 56.60 & 97.50 & / & / \\
RFT + High-RBF Data & 18.83 & 68.81 & 58.16 & 97.50 & / & / \\
\midrule
\multicolumn{7}{l}{\textit{Final Method (most iterations)}} \\
RFT + Noisy Data (5\%) + High-RBF & 23.23 & 66.62 & 67.46 & 100.00 & 32.23 & 72.43 \\
\bottomrule
\end{tabular}
\end{table}

\paragraph{Results and Analysis.}

Our experimental results in Table~\ref{tab:rft_results} lead to the following key conclusions:

$\bullet$ \textbf{Controlled noise injection enhances exploration.} Incorporating a small fraction (5\%) of noisy data during RFT improves Pass@k performance with minimal impact on Pass@1 accuracy. This suggests that preventing the model from becoming overly deterministic by exposing it to a controlled amount of negative samples helps it diversify its exploration.

$\bullet$ \textbf{Directly optimizing for exploration metrics is effective.} Data selection strategies based on high entropy and, particularly, high rollout branching factor yield the most significant improvements in Pass@k. This confirms that explicitly selecting for samples that exhibit high token-level diversity is a powerful method for enhancing a model's underlying exploration capabilities.

$\bullet$ \textbf{RFT with exploration-aware data selection surpasses strong RL baselines in Pass@k.} After several iterations, our RFT model trained with the noise injection and high rollout branching factor strategy outperforms the Qwen-2.5-32B-SimpleRL-Zoo baseline in Pass@k across all benchmarks, while its Pass@1 performance remains competitive. This demonstrates that our method effectively maintains and enhances the model's ability to find diverse correct solutions.

$\bullet$ \textbf{Enhanced exploration provides a foundation for long-term improvement.} While prioritizing exploration can introduce a slight, temporary trade-off in Pass@1 precision, the iterative nature of RFT allows the model to translate these exploratory gains into robust performance improvements over time. The substantial increase in Pass@k is particularly crucial, as it potentially provides a larger pool of successful trajectories, thereby offering significant room for further performance gains in complex reasoning tasks.
\begin{tcolorbox}
[colback=black!5!white,colframe=black!55!white,title=\textbf{\footnotesize \textsc{\ul{Takeaway for Maintaining Exploration Capabilities (Section 4.4.1)}}:}]
\begin{itemize}
[leftmargin=1mm]
\setlength\itemsep{0em}
    \item[\ding{224}] {\footnotesize 
    {\fontfamily{phv}\fontsize{8}{9}\selectfont
   In RFT, using data selection strategies that favor diversity—such as incorporating controlled noise or prioritizing samples with high rollout branching factor—can effectively maintain exploration capabilities and boost Pass@k performance.}}
\end{itemize}
\end{tcolorbox}

\subsubsection{Enhancing Optimization Efficacy: PPL- and Position-Based Advantage Shaping} % 这个要改一下

\begin{figure}[t]
  \centering

  % 图1
  \subfigure[Qwen2.5-7B\label{fig:subfig1}]{
    \includegraphics[width=0.43\textwidth]{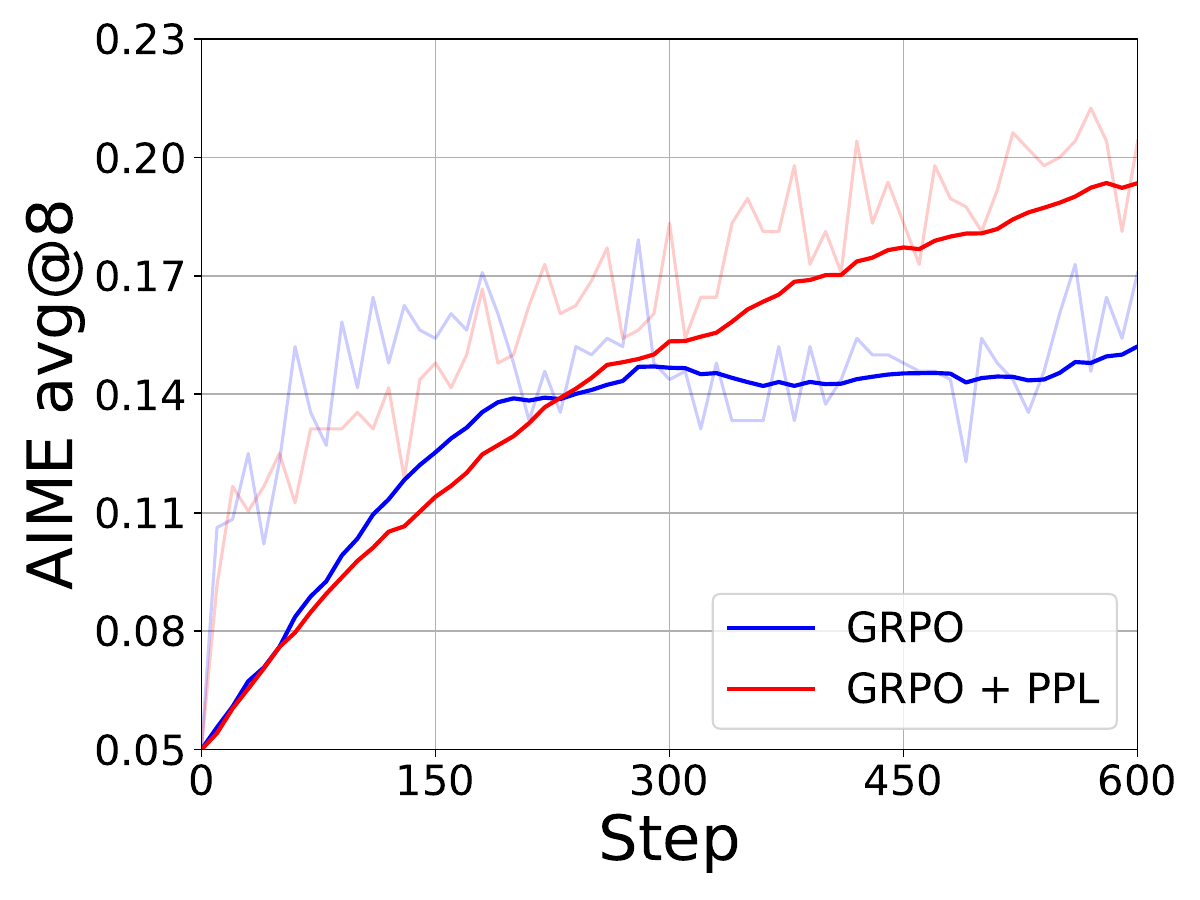}
  }
  \hfill
  % 图2
  \subfigure[Qwen2.5-Math-7B\label{fig:subfig2}]{
    \includegraphics[width=0.43\textwidth]{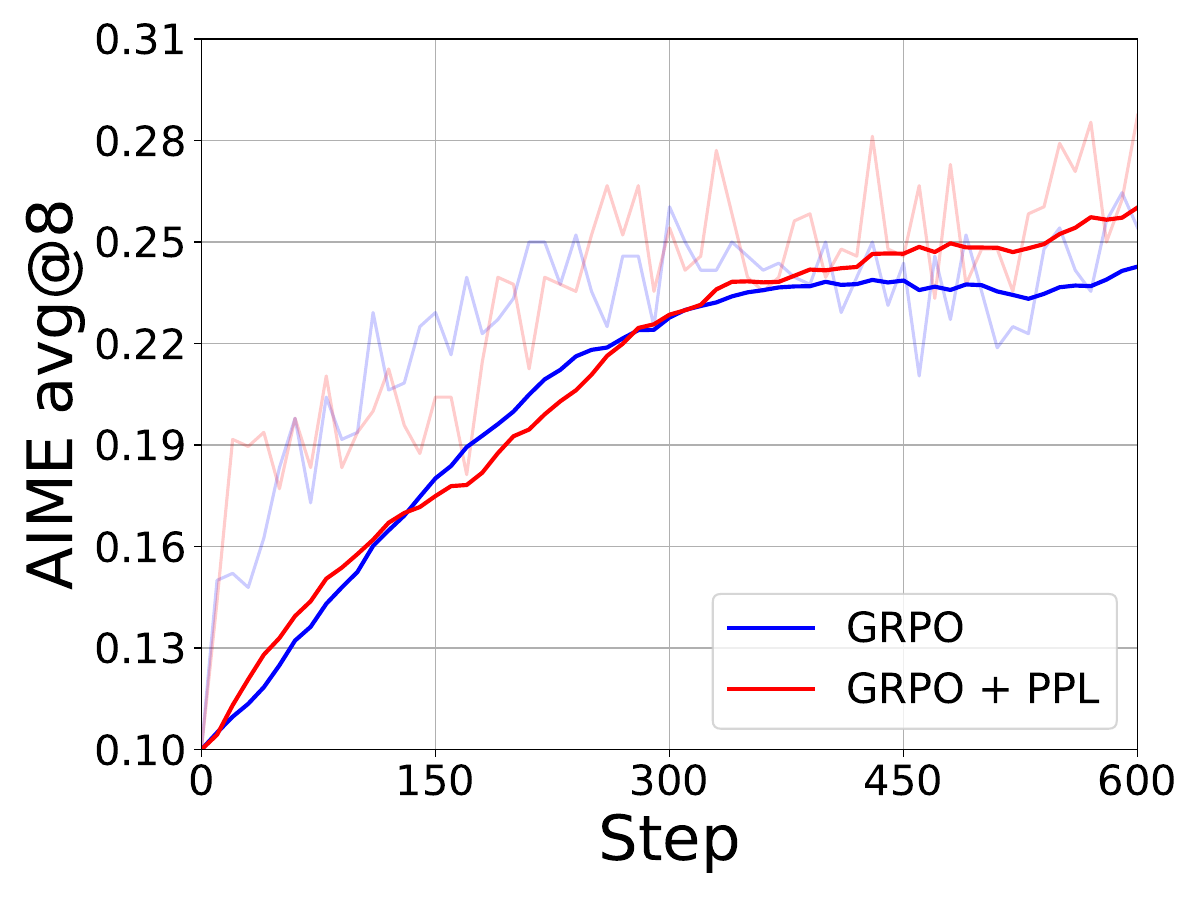}
  }
  \hfill
  % 图3
  \subfigure[Qwen2.5-7B\label{fig:subfig3}]{
    \includegraphics[width=0.43\textwidth]{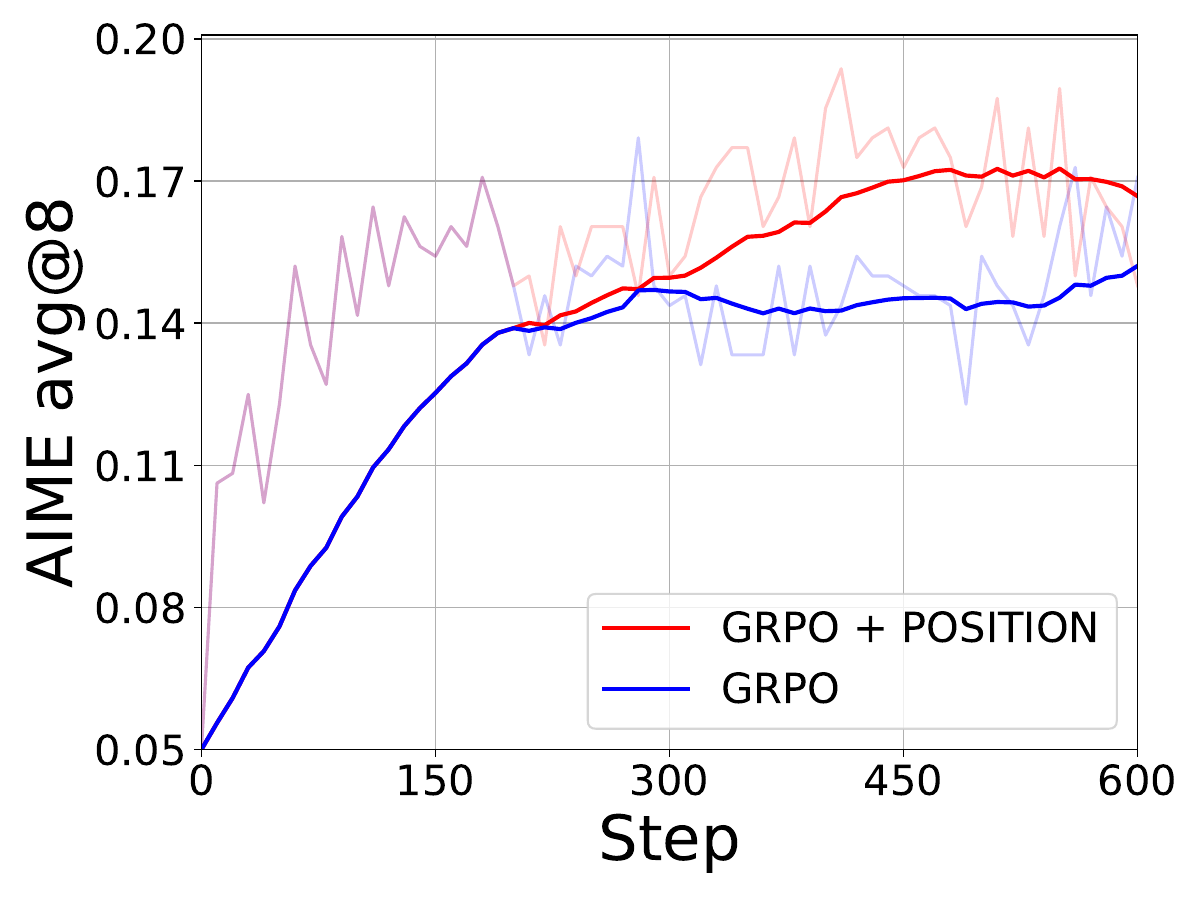}
  }
  \hfill
  % 图4
  \subfigure[Qwen2.5-Math-7B\label{fig:subfig4}]{
    \includegraphics[width=0.43\textwidth]{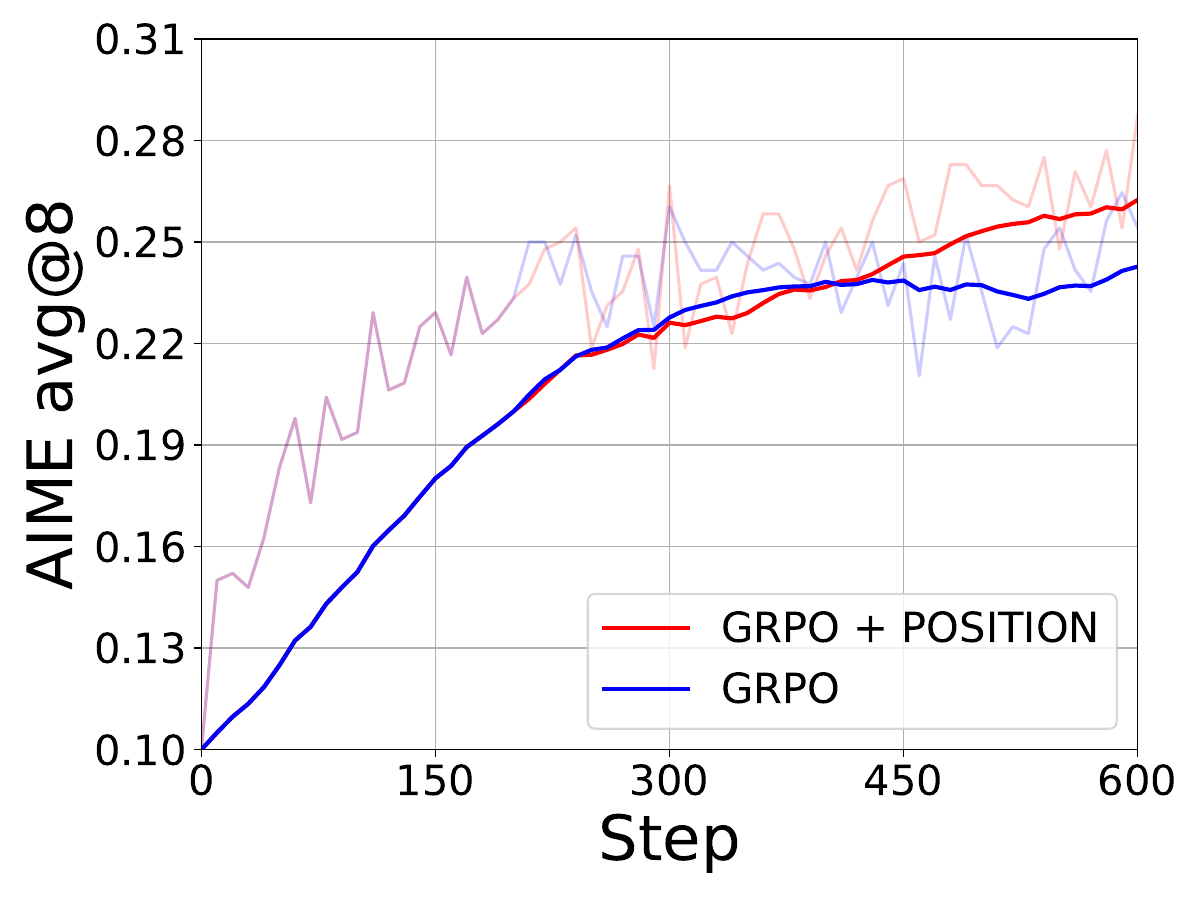}
  }

  \caption{Comparison of average accuracy change curves.}
  \label{fig:overall_acc}
\end{figure}

In Section~\ref{sec:inter}, we have found that perplexity (PPL) and position exert potential influences on reinforcement learning (RL) training. Here, we investigate how to leverage these insights to enhance the training of RLVR.

%Inspired by the findings mentioned in the previous section, we conduct the experiments on the two advantage shaping methods, which are also discussed in our previous work~\cite{}.
%Based on the internal information of the model, we shape and modify the advantage value estimation function in RLVR training, enabling the model to pay more attention to elements with greater learning potential during the RLVR process.

\begin{table*}[h]
\centering
\small
\setlength{\tabcolsep}{3pt}
\caption{Results on math benchmarks across The pass@k results is shown in~\ref{app:pk}. }
\begin{tabular}{lcccccccccc}
\toprule
\multicolumn{1}{c}{\multirow{2}{*}{\textbf{Method}}} & \multicolumn{2}{c}{\textbf{AIME24}} & \multicolumn{2}{c}{\textbf{AIME25}} & \multicolumn{2}{c}{\textbf{AMC23}} & \multicolumn{2}{c}{\textbf{MATH500}} & \multicolumn{2}{c}{\textbf{Avg.}} \\
\cmidrule(lr){2-3} \cmidrule(lr){4-5} \cmidrule(lr){6-7} \cmidrule(lr){8-9} \cmidrule(lr){10-11}
\multicolumn{1}{c}{} & avg@8 & maj@8 & avg@8 & maj@8 & avg@8 & maj@8 & avg@8 & maj@8 & avg@8 & maj@8 \\
\midrule
\multicolumn{11}{l}{\textit{Qwen2.5-7B}} \\
\cmidrule(lr){1-11}
BASE            & 7.50  & 9.88  & 1.67  & 1.56  & 38.44 & 52.50 & 58.63 & 75.40 & 26.56 & 34.84 \\
GRPO            & \underline{21.30} & \underline{24.17} & 15.40 & 15.54 & 59.38 & 65.00 & 80.83 & 84.80 & 44.23 & 47.38 \\
GRPO+PPL        & \textbf{22.08} & \textbf{25.03} &\textbf{ 18.75} & \textbf{20.24} & \underline{60.31} & \underline{67.50} & \textbf{82.90} & \textbf{86.00} & \textbf{46.01} & \underline{49.69} \\
GRPO+POSITION   & 20.00 & 21.40 & \underline{17.08} & \underline{17.68} & \textbf{63.44} & \textbf{75.00} & \underline{81.33} & \underline{85.20} & \underline{45.46} & \textbf{49.82} \\
\midrule
\multicolumn{11}{l}{\textit{Qwen2.5-Math-7B}} \\
\cmidrule(lr){1-11}
BASE           &15.42   &20.78  & 7.50 &13.38   &42.77 &52.47 &  57.60&67.41  &30.82  &38.51  \\
GRPO            & 27.08 & 31.25 & \underline{25.00} & \underline{25.85} & 67.81 & 72.50 & \underline{86.65} & \textbf{89.00} & 51.64 & 54.65 \\
GRPO+PPL        & \underline{31.25} & \underline{37.42} & \textbf{25.42} & \textbf{26.24} & \textbf{73.44} & \textbf{82.50} & \textbf{86.73} & \underline{88.80} & \textbf{54.21} & \textbf{58.74} \\
GRPO+POSITION   & \textbf{33.75} & \textbf{39.51} & 22.92 & 24.02 & \underline{71.56} & \underline{75.00} & 86.52 & 88.20 & \underline{53.69} & \underline{56.68} \\
\bottomrule
\end{tabular}
\label{tab:result-acc-highlighted}
\end{table*}

\paragraph{Methods.} In this part, we consider examining two simple adavantage shaping methods, described as follows: 

$\bullet$ \emph{PPL-based Advantage Shaping.}
As the first strategy, we adjust token advantages to favor low-PPL samples, where learning is concentrated. For each response $o^i$ in a batch, we compute its standardized log-PPL weight $w_{\text{ppl}}(o^i)$ using Eq.~\ref{eq:ppl_w}. The advantage \( A_t \) for each token \( t \) in that response is then modulated as follows:
\begin{equation}
\small
\tilde{A^i_t} = A^i_t \cdot \left(1 - \alpha \cdot w_{\text{ppl}}(o^i)\right).
\end{equation}
This method down-weights the updates from high-PPL samples, focusing the model's learning on more in-distribution reasoning paths.

$\bullet$ \emph{Position-based Advantage Shaping}
To focus optimization on the latter parts of reasoning sequences, we apply a position bonus to the token advantages. As motivated by our empirical analysis, we use the positional bonus $b^i_t$ defined in Eq.~\ref{eq:loc-bonus}. This bonus increases toward the end of the sequence and is applied based on the sign of the original advantage:
\begin{equation}
\small
\tilde{A}^{i'}_t = A^i_t + \mathrm{sign}(A^i_t) \cdot b^i_t.
\end{equation}
This approach encourages the model to allocate more learning effort toward the latter parts of its reasoning process.

\paragraph{Training Details.}For the PPL-based reward shaping method, we apply the advantage adjustment throughout the entire RLVR training process, as PPL's measure of the model's uncertainty over a sequence is consistently applicable across the entire training period. We set the scaling hyperparameter $\alpha=0.01$.
For the positional reward shaping method, as shown in Fig.~\ref{fig:loc_entropy}, our empirical analysis reveals that applying a positional bonus can cause a rapid rise in entropy. Therefore, we apply this method selectively. The bonus is only applied during the plateau stage, beginning at step 200 and continuing for 100 steps. Also, we set a small scaling factor $\gamma=0.1$ to moderate the entropy increase. We set the bonus direction $d=1.0$. The token's relative position score $r^i_t$ is calculated as $r^i_t = m \cdot (l^i_t - n)$, where $l^i_t \in [0,1]$ is the token's relative position in the sequence, with scaling and shifting parameters $m=15$ and $n=-0.5$.
\begin{figure}[!htbp]
  \centering

  % 左图：PPL-based shaping
  \subfigure[PPL-based shaping.\label{fig:ent1}]{
    \includegraphics[width=0.43\textwidth]{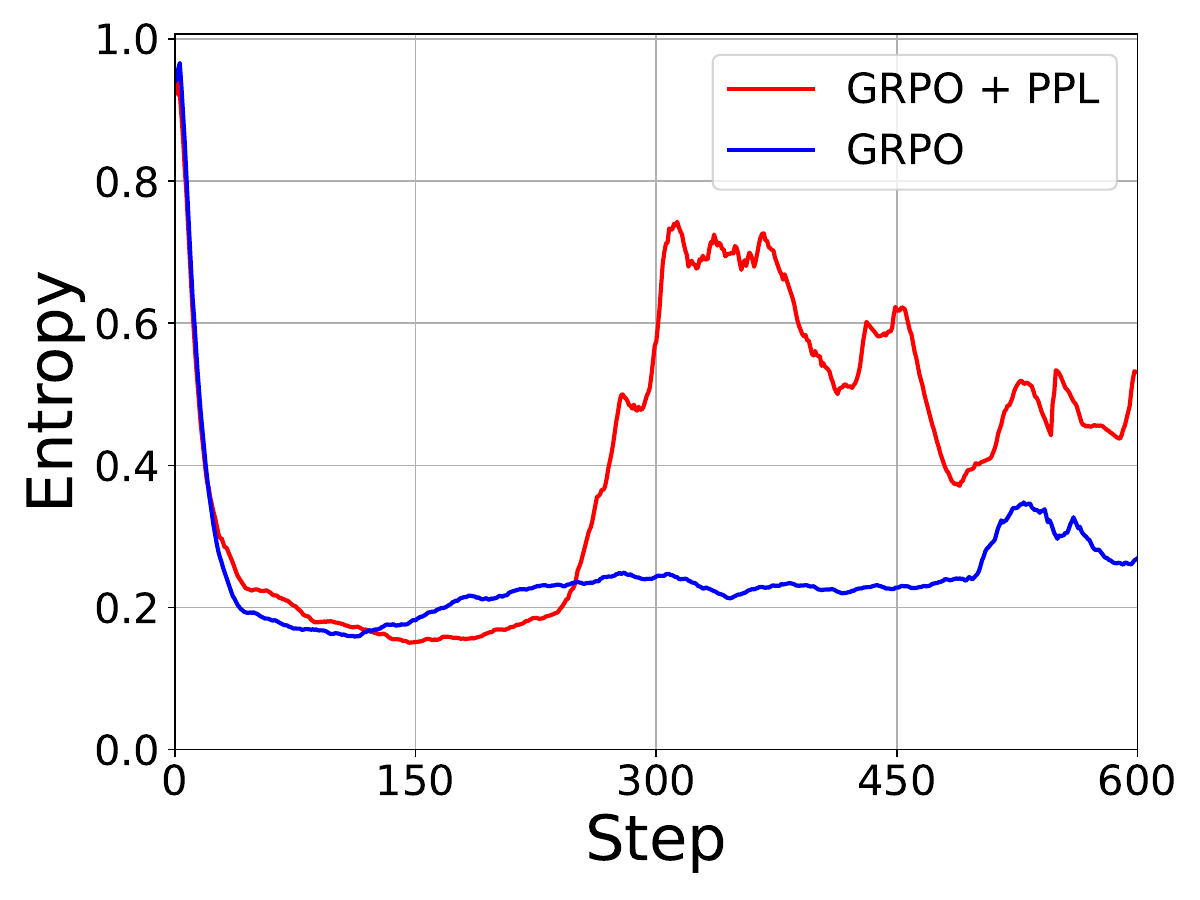}
  }
  \hspace{0.01\textwidth}
  % 右图：Position-based shaping
  \subfigure[Position-based shaping.\label{fig:ent2}]{
    \includegraphics[width=0.43\textwidth]{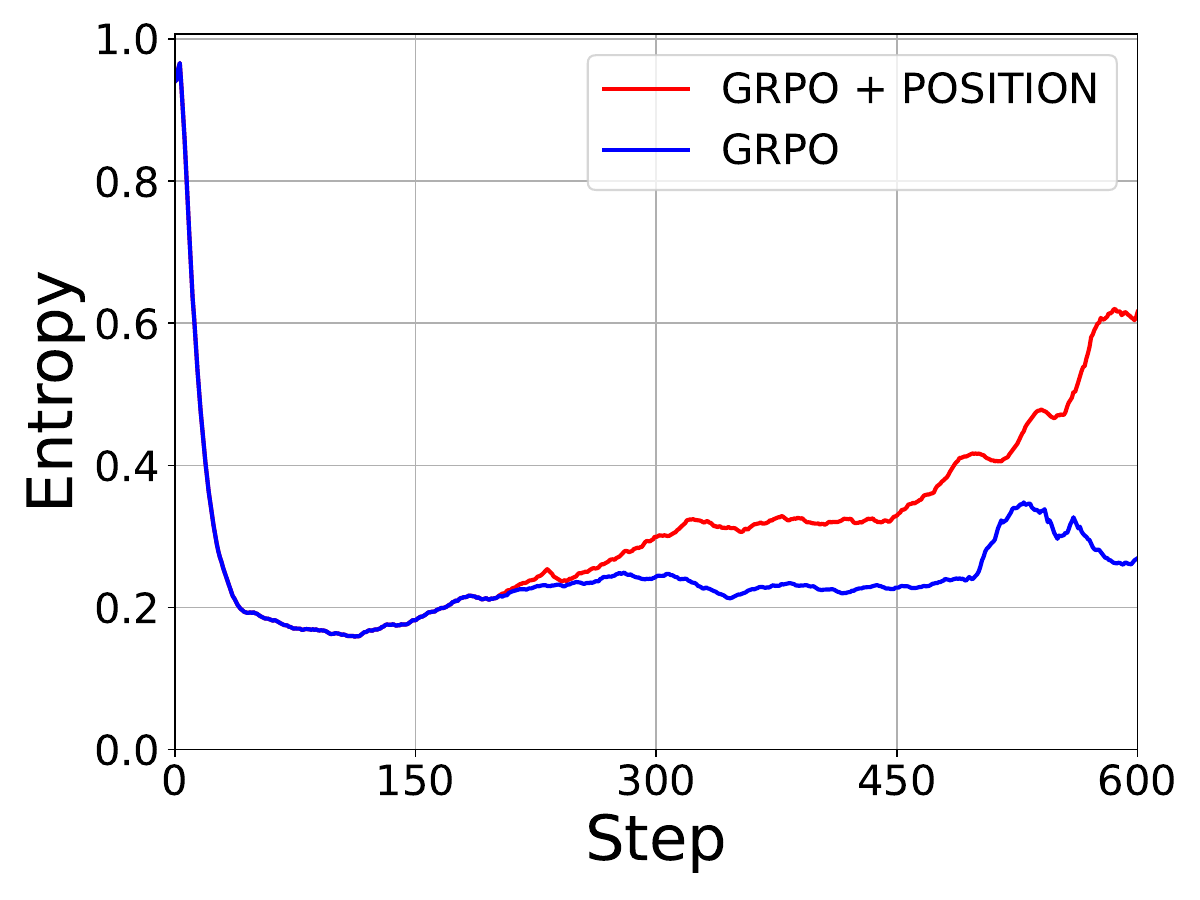}
  }

  \caption{Entropy dynamics for Qwen2.5-7B.}
  \label{fig:ppl-loc-entropy}
\end{figure}
\paragraph{Overall Performance.}
We evaluate our proposed methods on  mathematical reasoning benchmarks and analyze their impact on model behavior. As shown in Table~\ref{tab:result-acc-highlighted}, our approaches achieve subtantial improvements across the evaluation benchmarks. Compared to the GRPO baseline, they outperform it by an average of 1.51\% for the \texttt{Qwen2.5-7B} model and by 2.31\% for the \texttt{Qwen2.5-Math-7B} model, demonstrating the effectiveness of our targeted reward shaping. Moreover, our evaluations on GPQA and HumanEval reveal that both approaches exhibit enhanced generalization capabilities over the GRPO baseline.

\paragraph{Entropy Dynamics.}
%Our methods successfully maintain higher policy entropy during training compared to the GRPO baseline.
 As illustrated in Fig.~\ref{fig:ppl-loc-entropy}, our approaches sustain a higher level of entropy during the later stages of the plateau stage. 
 It exhibits a higher entroy trend compared to the GRPO baseline.
 This indicates that our method enables the model to retain substantial exploratory capability even in the later stages of training.
\begin{table}[H]
\centering
\caption{Comparison of average response length and token type counts in test set responses for Qwen2.5-7B.}
\label{tab:further}
 \resizebox{1.0\textwidth}{!}{%
\begin{tabular}{lcccc} 
\toprule
\textbf{Method} & \textbf{Mean Length} & \textbf{\parbox{3.5cm}{\centering Formal Reasoning Tokens}} & \textbf{\parbox{4cm}{\centering Logical Structuring Tokens}} & \textbf{\parbox{3cm}{\centering Metacognitive Reasoning Tokens}} \\
\midrule
GRPO            & 969.06        & 501.24   & 26.31  & 0.02  \\
GRPO+PPL        & 1841.44       & 1007.07  & 44.80  & 0.18  \\
GRPO+POSITION   & 1121.28 & 607.625  & 38.10  & 0.04  \\
\bottomrule
\label{tab:further}
\end{tabular}%
}
\end{table}
\paragraph{Response Pattern Analysis.}
We further analyze changes in response patterns by quantifying the distribution of token categories across all test sets. As shown in Table~\ref{tab:further}, both methods result in longer responses compared to the baseline, with a notable increase in tokens related to formal reasoning and logic. Formal reasoning tokens show the most significant increase, while the other categories, particularly metacognitive reasoning tokens, see smaller gains. This suggests that improving advanced cognitive abilities is inherently more difficult and may require more training steps. Case studies in Appendix~\ref{sec:case_stu} further reveal that both methods yield more detailed step-by-step breakdowns and a deeper display of the computational process compared to the baseline. Notably, the positional method encourages the model to attempt and backtrack from erroneous approaches, indicating a deeper reasoning process.

\begin{tcolorbox}
[colback=black!5!white,colframe=black!55!white,title=\textbf{\footnotesize \textsc{\ul{Takeaway for Enhancing Optimization Efficacy (Section 4.4.2)}}:}]
\begin{itemize}
[leftmargin=1mm]
\setlength\itemsep{0em}
    \item[\ding{224}] {\footnotesize 
    {\fontfamily{phv}\fontsize{8}{9}\selectfont
   The optimization efficiency can be consistently improved by leveraging the advantages of low-PPL instances and tokens appearing in later positions.}} 
\end{itemize}
\end{tcolorbox}
\section{Conclusion}

In this report, we present a systematic investigation of exploration mechanisms in LLMs under RL with verifiable rewards (RLVR). Our methodology combines a comprehensive literature review with original empirical analysis, focusing on four key dimensions: (1) exploration space shaping,  (2) entropy-performance interplay, and (3) RL performance optimization. The major findings of our study include:

$\bullet$ For exploration space shaping, we introduce two new metrics: $k$-rollout unsolvable problems and rollout branching factor—to augment conventional evaluation measures (pass@k and policy distribution entropy). Additionally, we propose to use the Venn diagram for comparative analysis of LLMs' reasoning capacities across different exploration regimes.
We observe that the sets of unsolvable problems for different methods, when compared to the base model, do not exhibit complete inclusion. This confirms that training does not simply expand the ability boundary but reconfigures it. Furthermore, tool augmentation can effectively expand the capacity boundary of large language models.

%We find that the sets of solvable problems from different methods against the base model show no complete inclusion, and tool augmentation can effectively extend the capactity boundary of LLMs.

%$\bullet$  Through systematic comparison of SFT and RL in RLVR, we demonstrate their contrasting impacts on exploration space dynamics, consistent with prior studies: RL narrows the exploration space of base models, while SFT can expand this space when provided with sufficient high-quality demonstration data. Beyond previous findings, our work reveals that RL can generate novel solutions exceeding the base model's original capacity boundaries.

$\bullet$ Through a detailed analysis of entropy-performance interplay in RLVR, we identify three core phenomena: stage-level dynamics, instance-level efficiency, and token-level significance. Our findings show that during the rising stage, models establish formal reasoning patterns primarily through entropy reduction in negative samples. Furthermore, plateau-stage analysis demonstrates that significant entropy changes occur predominantly in low-PPL responses and are concentrated in later-position tokens that contribute most to final decisions. 

$\bullet$ We consider improving RL performance improvement through two main aspects: (1) expanding exploration capacities and (2) enhancing the performance conversion efficacy. For exploration capacity enhancement, we systematically summarize existing methods and empirically validate several data selection strategies for RFT, confirming that prioritizing sample diversity is an effective method for enhancing exploration. Regarding performance conversion, we mainly review entropy-based optimization techniques and introduce two simple yet effective reward shaping methods that leverage instance perplexity and token position information.

RLVR offers a promising approach for enhancing LLMs' reasoning capabilities, while it also inherits limitations from conventional RL methods—such as training instability and low sample efficiency. We argue that a deeper understanding of RLVR algorithms is crucial for developing more capable reasoning models. Our work establishes a foundational framework that integrates existing insights with new empirical analysis, and we aim to further explore LLM exploration mechanisms in future research. 

%Looking ahead, future research should explore when and how to expand LLMs' reasoning boundaries. For frontier models lacking stronger teachers or human experts, active boundary expansion is essential for achieving superintelligence. Promising strategies include perturbing training distributions to elicit novel behaviors, incorporating new knowledge via distillation or SFT-RL hybrids, and exploring multi-modal learning pathways to push reasoning capabilities further.

\bibliographystyle{unsrt}
\bibliography{main.bib}

\appendix

\section{Appendix}

\subsection{Gradient Derivation}\label{app:gradient_derivation}

We derive the gradient of the GRPO objective $J_{\text{GRPO}}$ with respect to the logits $\mathbf{z} \in \mathbb{R}^V$. Recall the policy probability for token $o_t^i$:  
\[
\bm{\pi}_{\theta}(o_t^i) = \text{Softmax}(\mathbf{z})_i = \frac{e^{z_i}}{\sum_{j=1}^{V} e^{z_j}},
\]  
where $V$ is the vocabulary size. The gradient of $\bm{\pi}_{\theta}(o_t^i)$ w.r.t. $z_k$ is:  
\[
\frac{\partial \bm{\pi}_{\theta}(o_t^i)}{\partial z_k} = \bm{\pi}_{\theta}(o_t^i) \left( \mathbb{I}(o_t^i = v_k) - \bm{\pi}_{\theta}(v_k) \right),
\]  
with $\mathbb{I}(\cdot)$ the indicator function and $v_k$ the $k$-th vocabulary token. Applying the chain rule to $J_{\text{GRPO}}$:  
% \begin{align*}
% \frac{\partial J_{\text{GRPO}}}{\partial z_k} 
% &= \underbrace{\hat{r}_t \cdot \min\left( \hat{A}_t, \text{clip}(\hat{A}_t, 1-\epsilon, 1+\epsilon) \right)}_{\alpha_t} \cdot \frac{1}{\bm{\pi}_{\theta}(o_t^i)} \cdot \frac{\partial \bm{\pi}_{\theta}(o_t^i)}{\partial z_k} \\
% &= \alpha_t \left( \mathbb{I}(o_t^i = v_k) - \bm{\pi}_{\theta}(v_k) \right).
% \end{align*}
\begin{align*}
\frac{\partial J_{\text{GRPO}}}{\partial z_k}
& = \left[ \hat{r}_t \cdot \min\left( \hat{A}_t, \text{clip}(\hat{A}_t, 1-\epsilon, 1+\epsilon) \right) \right] \\
& \quad \cdot \frac{1}{\bm{\pi}_{\theta}(o_t^i)} \cdot \frac{\partial \bm{\pi}_{\theta}(o_t^i)}{\partial z_k} \\
&= \alpha_t \left( \mathbb{I}(o_t^i = v_k) - \bm{\pi}_{\theta}(v_k) \right).
\end{align*}
Vectorizing over the vocabulary $V$, the gradient is:  
\begin{equation}
\frac{\partial J_{\text{GRPO}}}{\partial \mathbf{z}} = \alpha_t \left( \bm{e}(o_t) - \bm{\pi}_{\theta} \right),
\label{eq:grad_logits}
\end{equation}  
where $\bm{e}(o_t) \in \mathbb{R}^V$ is the one-hot vector for token $o_t$, $\bm{\pi}_{\theta} \in \mathbb{R}^V$ is the policy distribution, and $\alpha_t = \hat{r}_t \cdot \min(\hat{A}_t, \text{clip}(\hat{A}_t, 1-\epsilon, 1+\epsilon))$.  

Crucially, the policy update operates on the language model head weights $\mathbf{W} \in \mathbb{R}^{V \times d}$, where $\mathbf{z} = \mathbf{W} \bm{h}$ and $\bm{h} \in \mathbb{R}^d$ is the last transformer layer's output. By the chain rule:  
\[
\frac{\partial J_{\text{GRPO}}}{\partial \mathbf{W}} = \frac{\partial J_{\text{GRPO}}}{\partial \mathbf{z}} \cdot \frac{\partial \mathbf{z}}{\partial \mathbf{W}} = \underbrace{\alpha_t \left( \bm{e}(o_t) - \bm{\pi}_{\theta} \right)}_{\in \mathbb{R}^V} \cdot \bm{h}^\top,
\]  
yielding a gradient matrix $\frac{\partial J_{\text{GRPO}}}{\partial \mathbf{W}} \in \mathbb{R}^{V \times d}$. The magnitude of this update is quantified by its Frobenius norm:  
\begin{equation}
\small
G_t = \left\| \alpha_t \left( \bm{e}(o_t) - \bm{\pi}_{\theta} \right) \bm{h}^\top \right\|_F,
\label{eq:grad_final}
\end{equation}  
where $\|\cdot\|_F$ denotes the Frobenius norm. This serves as the token-wise update magnitude proxy.

\subsection{Pass@k Results}\label{app:pk}
Results of pass@k on six benchmarks are shown in Tab.~\ref{tab:result-passk}. It can be seen that the average scores of our method on both the out-of-domain and in-domain benchmarks are higher than those of the GRPO baseline. However, all three methods struggle to surpass the performance of the base model on out-of-domain benchmarks, suggesting that applying reinforcement learning in the mathematics domain alone may weaken capabilities in other fields.
\begin{table*}[t]
\centering
\caption{Results for pass@k. All values are pass@8, except for HumanEval, which is pass@4.}
\label{tab:result-passk}
\resizebox{0.9\textwidth}{!}{%
\begin{tabular}{lccccccc}
\toprule
\textbf{Method} & \textbf{GPQA} & \textbf{HumanEval} & \textbf{AIME24} & \textbf{AIME25} & \textbf{AMC23} & \textbf{MATH500} & \textbf{Avg.} \\
\midrule
\multicolumn{8}{l}{\textit{Qwen2.5-7B}} \\
\cmidrule(lr){1-8}
BASE            & \textbf{63.13} & 14.18          & 20.30          & 7.32           & 77.50          & 87.00          & 44.91 \\
GRPO            & 48.98          & \underline{22.10} & \textbf{35.91} & \textbf{27.50} & 70.00          & \underline{90.00} & 49.08 \\
GRPO+PPL        & 50.50          & 21.49          & \underline{35.48} & 24.35          & \textbf{85.00} & \textbf{92.40} & 51.54 \\
GRPO+LOCATION   & \underline{61.11} & \textbf{23.02} & 29.75          & \underline{25.36} & \underline{80.00} & \underline{90.00} & 51.54 \\
\midrule
\multicolumn{8}{l}{\textit{Qwen2.5-Math-7B}} \\
\cmidrule(lr){1-8}
BASE            & \textbf{66.67} & \underline{26.68} & 33.76          & 21.61          & 70.96          & 84.46          & 50.69 \\
GRPO            & \underline{55.55} & 26.37          & 40.69          & \textbf{37.44} & \underline{85.00} & \underline{93.60} & 56.44 \\
GRPO+PPL        & 54.55          & \textbf{33.38} & \textbf{56.89} & \underline{33.18} & \textbf{87.50} & \textbf{94.40} & 59.98 \\
GRPO+LOCATION   & \underline{55.55} & 26.07          & \underline{55.33} & 30.85          & \underline{85.00} & 93.20          & 57.67 \\
\bottomrule
\end{tabular}%
}
\end{table*}
% \begin{table*}[htbp]
% \centering
% \begin{tabular}{|p{\textwidth}|}
% \hline
% \textbf{Prompt for Detecting Irrelevant Content} \\
% \hline
% Given a question and its corresponding answer, you are tasked with identifying if the response has a content redundancy issue. \textbf{Do not judge whether the answer is correct or incorrect.} Focus only on evaluating the content based on this criterion: \\
% \begin{itemize}
%     \item \textbf{Content redundancy}: The answer contains excessive repetition of meaningless words, or a large amount of irrelevant content, or gibberish.
% \end{itemize}
% Please evaluate the answer based on the criterion above. If the answer contains content redundancy, return \texttt{yes}. If it does not, return \texttt{no}. Enclose your answer in boxed.\\
% \hline
% \end{tabular}
% \caption{Prompt for detecting irrelevant content in responses.}
% \label{tab:prompt}
% \end{table*}
\subsection{Case Study}\label{sec:case_stu}
We compared the answers to the same question from models trained using three different methods: GRPO, GRPO+PPL, and GRPO+POSITION. The results are presented in Tab.~\ref{tab:grpo},Tab.~\ref{tab:ppl}, and Tab.~\ref{tab:positon} respectively. We found that the responses from the GRPO+PPL and GRPO+POSITION models were noticeably more granular, with more detailed formula derivations, making them significantly easier to understand than those from the GRPO model.
\begin{table*}[htbp]  % 这里的 [htbp] 是建议的浮动位置，表示尽量在此处显示
\centering  % 使表格居中
\caption{Answer from GRPO.}
\label{tab:grpo}
\begin{tabular}{|p{0.95\textwidth}|} % 使用p{宽度}来控制列宽并自动换行
\hline
Question \\ \hline
Every morning Aya goes for a $9$-kilometer-long walk and stops at a coffee shop afterwards. When she walks at a constant speed of $s$ kilometers per hour, the walk takes her 4 hours, including $t$ minutes spent in the coffee shop. When she walks $s+2$ kilometers per hour, the walk takes her 2 hours and 24 minutes, including $t$ minutes spent in the coffee shop. Suppose Aya walks at $s+\frac{1}{2}$ kilometers per hour. Find the number of minutes the walk takes her, including the $t$ minutes spent in the coffee shop. \\ \hline
Answer from GRPO \\ \hline
To determine the number of minutes the walk takes Aya when she walks at a speed of \( s + \frac{1}{2} \) kilometers per hour, we need to first set up the problem using the information given in the problem. Let's start by defining the variables and the relationships involved.The total distance of the walk is 9 kilometers, and Aya spends \( t \) minutes in the coffee shop. The total time for the walk at a speed of \( s \) kilometers per hour takes 4 hours. This total time includes both the time taken to walk and the time spent in the coffee shop. Therefore, the time taken to walk 9 kilometers at a speed of \( s \) kilometers per hour is given by:\[\text{Time taken to walk} = \frac{9}{s} \text{ hours}\]The total time taken for the walk including the time spent in the coffee shop is 4 hours. So, we have:\[\text{Time taken to walk} + \text{Time spent in coffee shop} = 4 \text{ hours}\]Therefore,\[\frac{9}{s} + \frac{t}{60} = 4 \quad \text{(1)}\]For the second scenario, where Aya walks at a speed of \( s + 2 \) kilometers per hour, the total time taken for the walk including the time spent in the coffee shop is 2 hours and 24 minutes, which is equivalent to 2.4 hours. So we have:\[\frac{9}{s + 2} + \frac{t}{60} = 2.4 \quad \text{(2)}\]Now we have two equations (1) and (2): ...\\Subtracting equation (2) from equation (1), we get:...\\
Simplifying the left side of the equation, we have:...\\Combining the fractions on the left side:...\\Multiplying both sides by \( s(s + 2) \):...\\Dividing both sides by 1.6:...\\So, we have the quadratic equation:...\\We solve this quadratic equation using the quadratic formula \( s = \frac{-b \pm \sqrt{b^2 - 4ac}}{2a} \), where \( a = 1 \), \( b = 2 \), and \( c = -11.25 \):...\\This gives two solutions:...\\Since the speed \( s \) must be positive, we discard the negative solution. Therefore,\[s = 2.5 \text{ km/h}\]
...\\Therefore, the number of minutes the walk takes Aya, including the time spent in the coffee shop, when she walks at a speed of \( s + \frac{1}{2} \) km/h is \(\boxed{204}\). \\ \hline
\end{tabular}
\end{table*}

\begin{table*}[htbp]  % 这里的 [htbp] 是建议的浮动位置，表示尽量在此处显示

\caption{Answer from GRPO+PPL}
\label{tab:ppl}
\centering  % 使表格居中
\begin{tabular}{|p{0.95\textwidth}|} % 使用p{宽度}来控制列宽并自动换行
\hline
Question \\ \hline
Every morning Aya goes for a $9$-kilometer-long walk and stops at a coffee shop afterwards. When she walks at a constant speed of $s$ kilometers per hour, the walk takes her 4 hours, including $t$ minutes spent in the coffee shop. When she walks $s+2$ kilometers per hour, the walk takes her 2 hours and 24 minutes, including $t$ minutes spent in the coffee shop. Suppose Aya walks at $s+\frac{1}{2}$ kilometers per hour. Find the number of minutes the walk takes her, including the $t$ minutes spent in the coffee shop. \\ \hline
Answer from GRPO+PPL \\ \hline
...\\**Step 1: Analyze the first scenario where Aya walks at a speed of \(s\) kilometers per hour.**- **Walking Time:** When Aya walks at a constant speed of $s$ kilometers per hour, the time she spends walking is the distance divided by her speed. The distance she walks is 9 kilometers.  - The walking time can be calculated as: ...\\ - **Total Time:** The total time for the walk, including the time spent in the coffee shop, is 4 hours.  - The time spent in the coffee shop is $t$ minutes, which we convert to hours by dividing by 60:    $t$ minutes = $\frac{t}{60}$ hours.  - Thus, the total time is the sum of the walking time and the time spent in the coffee shop, which equals 4 hours.  Thus,...\\Substitute the expression for the walking time into this equation:...\\Multiply through by 60 to clear the fractions:...\\**Step 2: Analyze the second scenario where Aya walks at a speed of \(s + 2\) kilometers per hour.**- **Walking Time:** When Aya walks at a speed of \(s + 2\) kilometers per hour, the walking time is the distance divided by this new speed, which is $\frac{9}{s+2}$ hours.- **Total Time:** The total time for the walk, including the time spent in the coffee shop, is 2 hours and 24 minutes. Convert 2 hours and 24 minutes to hours.  - 2 hours and 24 minutes is equal to \(2 + \frac{24}{60} = 2 + 0.4 = 2.4\) hours.  - Thus, the total time is the sum of the walking time and the time spent in the coffee shop, which equals 2.4 hours.  Thus,...\\Substitute the expression for the walking time into this equation:...\\Multiply through by 60 to clear the fractions:...\\Now we have two equations:1. $\frac{540}{s} + t = 240$2. $\frac{540}{s+2} + t = 144$Let's subtract equation (2) from equation (1):...\\Simplify the left side:...\\Combine the fractions:...\\Thus,...\\Now we solve the equation:...\\Multiply both sides by $s(s+2)$ to clear the fraction:...\\Divide both sides by 96:...\\Thus, we have:...\\This leads to the quadratic equation:...\\Multiply through by 4 to clear the fraction:...\\Rearrange the terms:...\\Now we solve the quadratic equation $4s^2 + 8s - 45 = 0$ for $s$ using the quadratic formula $s = \frac{-b \pm \sqrt{b^2 - 4ac}}{2a}$ where $a = 4$, $b = 8$, and $c = -45$.The quadratic formula is:...\\Substitute the coefficients into the formula:...\\Calculate the discriminant:...\\Thus,...\\Since $\sqrt{784} = 28$, we have two possible solutions for $s$:...\\Since speed cannot be negative, we discard $s = -\frac{9}{2}$. Thus,...\\Now that we have the value of \( s \), we substitute \( s = \frac{5}{2} \) back into one of the original equations to find \( t \). We will use Eq. (1):...\\Substitute \( s = \frac{5}{2} \):...\\Calculate \(\frac{540}{\frac{5}{2}}\):...\\Thus,...\\Solve for \( t \):...\\So, \( t = 24 \) minutes.**Step 3: Determine the time it takes Aya to complete the walk if she walks at a speed of \( s + \frac{1}{2} \) kilometers per hour.**First, we need to determine the speed \( s + \frac{1}{2} \).\( s = \frac{5}{2} \)
...\\So, the total time in minutes is:\(\boxed{204}\) \\ \hline
\end{tabular}
\end{table*}
\begin{table*}[htbp]  % 这里的 [htbp] 是建议的浮动位置，表示尽量在此处显示
\centering  % 使表格居中
\caption{Answer from GRPO+POSITION.}
\label{tab:positon}
\begin{tabular}{|p{0.95\textwidth}|} % 使用p{宽度}来控制列宽并自动换行
\hline
Question \\ \hline
Every morning Aya goes for a $9$-kilometer-long walk and stops at a coffee shop afterwards. When she walks at a constant speed of $s$ kilometers per hour, the walk takes her 4 hours, including $t$ minutes spent in the coffee shop. When she walks $s+2$ kilometers per hour, the walk takes her 2 hours and 24 minutes, including $t$ minutes spent in the coffee shop. Suppose Aya walks at $s+\frac{1}{2}$ kilometers per hour. Find the number of minutes the walk takes her, including the $t$ minutes spent in the coffee shop. \\ \hline
Answer from GRPO+Location \\ \hline
To find the value of \(s\), \(t\) and then the time Aya takes when walking at \(s + \frac{1}{2}\) kilometers per hour, we start by setting up equations based on the given information about Aya's walking speed and the total time for her walk including the time at the coffee shop.First, consider the case where Aya walks at a constant speed of \(s\) kilometers per hour. The total distance she walks is 9 kilometers and the total time for her walk including the \(t\) minutes at the coffee shop is 4 hours. The walking time at speed \(s\) can be written as \(\frac{9}{s}\) hours. Since the total time for the walk and the coffee shop visit is 4 hours, the walking time is \(4 - \frac{t}{60}\) hours. So we have the equation for the walking time at speed \(s\):\[\frac{9}{s} = 4 - \frac{t}{60}\]Next, consider the case where Aya walks at a constant speed of \(s+2\) kilometers per hour. The total time for her walk including the \(t\) minutes at the coffee shop is 2 hours and 24 minutes, which is \(2 + \frac{24}{60} = 2.4\) hours. The walking time at speed \(s+2\) can be written as \(\frac{9}{s+2}\) hours. Since the total time for the walk and the coffee shop visit is 2.4 hours, the walking time is \(2.4 - \frac{t}{60}\) hours. So we have the equation for the walking time at speed \(s+2\):...\\
Now we have a system of two equations:...\\We can now set these two expressions for \(4 - \frac{t}{60}\) and \(2.4 - \frac{t}{60}\) equal to each other:...\\ Rearrange the equation to get:...\\This equation seems problematic, so let's solve the system of equations for \(s\) and \(t\) first and then find the value of the walking time when Aya walks at \(s + \frac{1}{2}\) kilometers per hour. From the first equation, we have:...\\From the second equation, we have:...\\Now we have two equations in terms of \(s\) and \(t\):...\\We can now set the right sides of these two equations equal to each other:...\\\textbf{This simplifies to \(4 = 2.4\), which is clearly false}. So we need to solve these two equations for \(s\) and \(t\) separately. From the first equation:...\\Multiply both sides by 60 to clear the fractions:...\\Rearrange to get:...\\From the second equation:...\\Multiply both sides by 60 to clear the fractions:...\\Now we have the system of equations:...\\We now have two equations in terms of \(s\) and \(t\). We can solve these two equations for \(s\) and \(t\).First, we can equate the two expressions for \(t\) from these two equations. From equation \((1)\):...\\From equation \((2)\):...\\Now we have an equation in terms of \(s\):...\\Rearrange terms to combine the fractions:...\\Simplify the left side of the equation:...\\Simplify the expression in the numerator of the right side:...\\Now we can cross multiply to obtain:...\\Simplify the equation:...\\So now we have the equation:...\\Rearrange terms to form a quadratic equation:...\\We now have a quadratic equation to solve for \(s\). We can use the quadratic formula \(s= \frac{-b\pm \sqrt{b^2-4ac}}{2a}\), where \(a=4\), \(b=8\), and \(c= -45\). First compute the discriminant:...\\So \(\sqrt{b^2- 4ac} = \sqrt{784} = 28\). Now we can apply the formula for the roots of the equation:...\\This gives two potential solutions for \(s\):...\\
....\\Thus the number of minutes the walk takes her including \( t\) minutes spent in the coffee shop when she walks at \(s +\frac{1}{2}\) kilometers per hour is \(\boxed{204}\). \\ \hline
\end{tabular}
\end{table*}

% \end{document}

\end{document}